\documentclass{article}

\usepackage{iclr2025_conference,times}
\IfFileExists{math_commands.tex}{%%%%% NEW MATH DEFINITIONS %%%%%

\usepackage{amsmath,amsfonts,bm}

\def\eqref#1{equation~\ref{#1}}
\def\1{\bm{1}}

\DeclareMathAlphabet{\mathsfit}{\encodingdefault}{\sfdefault}{m}{sl}
\SetMathAlphabet{\mathsfit}{bold}{\encodingdefault}{\sfdefault}{bx}{n}

}{}
\usepackage{hyperref}
\usepackage{url}
\usepackage{amsmath,amssymb,bm,mathtools}
\usepackage{algorithm}
\usepackage[noend]{algpseudocode}
\usepackage{booktabs}
\usepackage{graphicx}
\usepackage{subcaption}
\usepackage{multirow}
\usepackage{makecell}
\usepackage{array}
\usepackage{microtype}
\usepackage{enumitem}
\usepackage{xspace}
\usepackage{tabularx}
\usepackage[table]{xcolor}

\definecolor{tablegray}{RGB}{247,247,247}
\definecolor{caopdblue}{RGB}{229,240,250}

\newcolumntype{Y}{>{\centering\arraybackslash}X}

\newcommand{\todofig}[1]{\textcolor{red}{\textbf{[TODO figure #1]}}}

\newcommand{\ph}{--}
\newcommand{\placeholderbox}[2]{%
\fbox{\parbox[c][#1][c]{0.94\linewidth}{\centering\textbf{Placeholder figure}\\[4pt]#2}}}

\title{
    CA-OPD: Confidence-Aware On-Policy \\ Distillation
    for Structured Visual Prediction
}

\author{
\textbf{Menghao Li}\textsuperscript{1,*} \quad
\textbf{Linjie Mu}\textsuperscript{2,*} \quad
\textbf{Haotian Hu}\textsuperscript{3,*,\S} \quad
\textbf{Yannian Gu}\textsuperscript{2} \quad
\\[2pt]
\textbf{Lujiayi Xue}\textsuperscript{4} \quad
\textbf{Liujian Tang}\textsuperscript{3} \quad
\textbf{Yu Zhang}\textsuperscript{3} \quad
\textbf{Fanyi Wang}\textsuperscript{3,$\dagger$}
\\[5pt]
{\small
\textsuperscript{1} Tianjin University
\qquad
\textsuperscript{2} Shanghai Jiao Tong University
}
\\[1pt]
{\small
\textsuperscript{3} StepX
\qquad
\textsuperscript{4} University of Science and Technology of China
}
}

\iclrfinalcopy

\begin{document}

\maketitle

% \begingroup
% \renewcommand{\thefootnote}{\fnsymbol{footnote}}
% \footnotetext[1]{Equal contribution.}
% \footnotetext[2]{Corresponding author.}
% \footnotetext[4]{Project Leader.}
% \endgroup

\setcounter{footnote}{0}

\begin{abstract}
Autoregressive vision language models unify heterogeneous perception tasks but are highly susceptible to compounding errors. 
On-policy distillation (OPD) bridges the training-inference mismatch by training students on their own rollouts. 
However, unreliable student predictions, especially early in training, can derail the trajectory and degrade the quality of teacher supervision. 
While recent interleaved distillation methods allow the teacher to verify and replace student tokens, they primarily rely on rigid ranking metrics rather than exact teacher confidence, and they overlook how intervention decisions can inform token-level supervision. 
To address this, we introduce Confidence-Aware On-Policy Distillation (CA-OPD), a framework that couples reliable rollout construction with adaptive supervision. 
CA-OPD utilizes teacher confidence to selectively correct unreliable student transitions, gradually transferring rollout control to the student via a strict-to-relaxed schedule. 
Crucially, CA-OPD aligns knowledge transfer with these intervention decisions: corrected positions receive direct cross-entropy supervision from the teacher's prediction, while retained positions benefit from the teacher's full predictive distribution. 
Evaluated in a multi-teacher setting for GUI grounding and optical character
recognition, CA-OPD substantially improves the Qwen3.5-0.8B baseline across
all six target benchmarks, including gains of $9.50$ points on ScreenSpot-Pro
and $6.72$ points on OCRBench-v2 English.
Controlled studies further show that the gains depend on intervention
placement, progressive rollout control, and intervention-aligned supervision,
rather than intervention frequency alone.
\end{abstract}

\section{Introduction}
\label{secintro}

Vision language models (VLMs) increasingly solve heterogeneous perception tasks through a unified autoregressive interface~\citep{peng2023kosmos,wang2026stepx}. 
For instance, optical character recognition (OCR) outputs are generated as text, while visual grounding and graphical user interface (GUI) interaction serialize locations or actions as token sequences~\citep{kim2021donut,chen2021pix2seq,you2024ferret}. 
While this formulation provides a flexible, task-agnostic interface, it makes structured prediction inherently sequential: every generated token becomes part of the next input. 
Once an incorrect token is committed, subsequent predictions are conditioned on a corrupted prefix. 
A local error can therefore compound, degrading both the immediate prediction and the entire trajectory along which the remainder of the output is generated~\citep{ranzato2015sequence,kim2022ocr}.

This autoregressive dependency poses a significant challenge for knowledge distillation (KD). 
Traditional offline and sequence-level distillation train the student on stable, reference- or teacher-generated trajectories~\citep{hinton2015distilling,kim2016sequence}. 
However, during inference, the student is evaluated on prefixes produced by its own policy. 
On-policy distillation (OPD) mitigates this mismatch by allowing the student to generate the rollout, querying the teacher for supervision on the states the student actually visits~\citep{agarwal2024policy,gu2024minillm,yuan2026vision}. 
Yet, the very property that aligns OPD with inference also introduces a critical vulnerability. 
When the student is unreliable, especially early in training, an incorrect prediction can enter the prefix and derail all subsequent states where teacher supervision is applied~\citep{xu2025speculative}.

Recent interleaved distillation methods offer a natural remedy for this tension. 
Instead of blindly committing every student prediction to the rollout, the teacher can verify student proposals and intervene when necessary. 
Speculative Knowledge Distillation (SKD)~\citep{xu2025speculative}, for example, uses the teacher's ranking of a student proposal to decide whether the proposal should remain in the generated prefix. 
Such verification provides a useful middle ground between relying entirely on student generation during OPD and using trajectories dictated by the teacher.
It protects the rollout from severe student errors while retaining substantial exposure to states induced by the student policy.

Nevertheless, two limitations remain. 
First, rank does not directly measure how strongly the teacher supports a student proposal. 
A token can rank highly even when the teacher distribution is uncertain, whereas proposals with the same rank can receive substantially different probability mass. 
The decision to trust a student transition is therefore better informed by the exact teacher confidence assigned to that proposal. 
Second, teacher intervention contains useful learning signals beyond simply repairing the rollout. 
When the teacher rejects a student proposal and commits its own prediction, it identifies a concrete local mistake. 
Treating this event solely as a rollout-correction mechanism leaves this corrective signal underutilized.
Conversely, when the teacher already supports the student proposal, preserving the richer teacher distribution can provide informative supervision. 
These observations suggest that rollout control and token-level supervision should be jointly optimized.

\begin{figure*}[t]
\centering
\includegraphics[width=\textwidth]{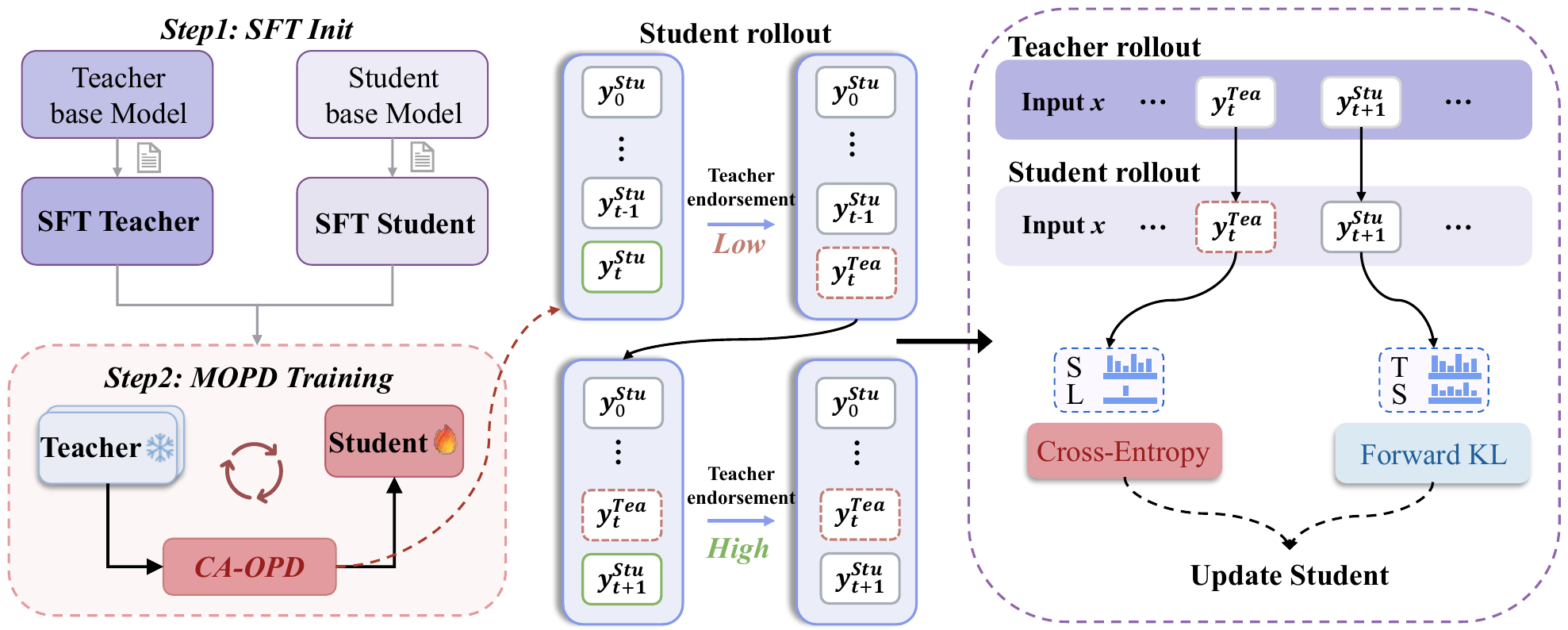}
\caption{
\textbf{Confidence-aware on-policy distillation for structured visual prediction.}
CA-OPD evaluates each student proposal using teacher confidence. Low-support proposals are replaced by teacher tokens and receive direct corrective supervision, while supported proposals remain in the rollout and retain distributional supervision. The mixed prefix is used for subsequent autoregressive generation.
}
\label{fig:intro_case}
\end{figure*}

To address this, we introduce CA-OPD, a confidence-aware on-policy distillation framework that couples reliable rollout construction with adaptive token-level supervision. 
CA-OPD leverages teacher confidence to assess student proposals and selectively correct unreliable transitions. 
Teacher intervention is stronger early in training when the student is less reliable and gradually decreases, allowing the rollout to smoothly move toward greater student control. 
Crucially, the outcome of this verification dictates how knowledge is transferred: \textit{corrected positions receive direct supervision from the teacher's prediction, while retained positions continue to benefit from the teacher's full predictive distribution.}
As illustrated in Figure~\ref{fig:intro_case}, CA-OPD preserves the on-policy character of training while selectively preventing poorly supported student transitions from being committed to the prefix. 
This same gate determines the supervision used at each position, elegantly connecting state visitation and knowledge transfer within a single autoregressive process.

We evaluate CA-OPD in a multi-teacher setting that distills complementary
GUI grounding and OCR capabilities into Qwen3.5-0.8B.
Relative to the Qwen3.5-0.8B baseline, CA-OPD improves all six target
benchmarks, with gains of $9.08$ points on ScreenSpot-v2, $9.50$ points on
ScreenSpot-Pro, $6.72$ points on OCRBench-v2 English, and $9.77$ points on
OCRBench-v2 Chinese.
It also outperforms Offline KD, standard OPD, and SKD on every target
benchmark, indicating that the gains cannot be attributed to distillation
alone.
Furthermore, controlled studies demonstrate that our gains cannot be explained by intervention frequency alone. Specifically, we show that (1) where the teacher intervenes matters, as randomly placed interventions provide minimal benefit compared to confidence-based ones at identical budgets; (2) a strict-to-relaxed schedule outperforms both fixed and reversed alternatives; and (3) aligning the supervision objective with the intervention decision materially contributes to the overall improvement.
Together, these results indicate that effective on-policy distillation depends not only on how often the teacher intervenes, but also on the strategic placement of interventions, the evolution of control over training, and the adaptive supervision of corrected versus retained positions.

Our contributions are summarized below:
\begin{itemize}[leftmargin=*,topsep=2pt,itemsep=2pt]
\item We identify a coupled challenge in on-policy distillation for structured autoregressive prediction: unreliable student transitions corrupt subsequent states conditioned for generation, and existing methods fail to utilize the teacher's intervention decision as a signal for token-level supervision.

\item We introduce CA-OPD, which uses teacher confidence to selectively repair student-generated prefixes, progressively transfers rollout control to the student, and applies intervention-aligned supervision: cross-entropy at replaced positions and distributional distillation at retained positions.

\item We demonstrate consistent gains across GUI grounding and OCR in a multi-teacher distillation setting while preserving held-out capabilities. Controlled studies confirm that improvements stem from strategic intervention placement, the direction of the rollout-control schedule, and adaptive supervision, rather than sheer intervention frequency.
\end{itemize}

\section{Related Work}
\label{secrelated}

\paragraph{Autoregressive structured visual prediction.}
Representing structured visual outputs as token sequences is a prevailing paradigm for unifying heterogeneous perception tasks.
For instance, Pix2Seq~\citep{chen2021pix2seq} and Donut~\citep{kim2021donut} serialize bounding boxes and document texts, while Unified-IO models map diverse multimodal outputs into a shared autoregressive interface~\citep{lu2022unified,lu2024unified}.
Recent VLMs similarly express visual grounding and GUI interactions through language-like coordinate sequences or action tokens~\citep{peng2023kosmos,chen2023shikra,you2024ferret,cheng2024seeclick,lin2025showui}.
While this formulation provides a flexible, task-agnostic interface, it makes structured prediction inherently sequential.
Because every committed token becomes part of the subsequent prefix, errors in OCR symbols, coordinates, or actions inevitably compound.
Such error propagation is particularly consequential for structured outputs, where a small local deviation can change the semantic interpretation of the entire prediction or redirect subsequent generation toward incorrect states.
% This property makes the reliability of intermediate prefixes an important consideration not only at inference time but also during training.
% Rather than proposing new visual representations, our work focuses on ensuring reliable knowledge transfer under this strict autoregressive dependency.

\paragraph{Knowledge distillation for autoregressive models.}
Traditional knowledge distillation (KD) transfers knowledge via teacher predictive distributions~\citep{hinton2015distilling}, while sequence-level KD relies on complete teacher-generated trajectories~\citep{kim2016sequence}.
Recent advances in language and multimodal model distillation primarily broaden what knowledge is transferred or which divergence is minimized.
For example, Distilling Step-by-Step leverages teacher rationales~\citep{hsieh2023distilling}, MiniLLM optimizes reverse-KL divergence to better match generation distributions~\citep{gu2024minillm}, and LLaVA-KD transfers complex multimodal relations~\citep{cai2025llavakd}.
While these approaches enrich the available supervision, their effectiveness is bottlenecked by the autoregressive states where this supervision is applied. Because errors compound in structured prediction, static teacher-generated prefixes fail to reflect true inference trajectories. CA-OPD complements these representation-focused methods by jointly optimizing rollout construction and token-level supervision.

\paragraph{On-policy and interleaved distillation.}
The distribution shift between training and inference states is a classic challenge in sequential prediction, historically addressed by techniques like scheduled sampling~\citep{bengio2015scheduled} and DAgger~\citep{ross2011reduction}.
In modern KD, Generalized Knowledge Distillation (GKD) addresses this by querying the teacher on student-generated rollouts, forcing the student to learn from its own inference-time errors~\citep{agarwal2024policy}.
SKD refines this via interleaved trajectories, where the teacher replaces student proposals that rank poorly under its distribution~\citep{xu2025speculative}.
Although such interleaving improves the reliability of student-generated trajectories, a rank-based criterion only measures the relative ordering of a proposal and does not directly quantify how strongly the teacher supports it.
% Moreover, applying a fixed intervention rule throughout optimization overlooks the fact that the reliability of the student policy changes substantially during training.
CA-OPD refines this interleaved approach by replacing rigid ranking with the teacher's confidence to guide interventions.
It dynamically relaxes this intervention threshold as the student improves, and explicitly couples rollout control with adaptive token-level supervision, thereby unifying state visitation and knowledge transfer.

\section{Problem Formulation}
\label{secproblem}

\subsection{Structured Autoregressive Prediction}

Let $x=(I,q)$ denote an input pair comprising an image $I$ and a textual instruction $q$. An autoregressive model generates an output sequence $y=(y_1,\ldots,y_L)$ over a vocabulary $\mathcal{V}$ such that:
\begin{equation}
\pi(y\mid x)
=
\prod_{t=1}^{L}
\pi(y_t\mid x,y_{<t}).
\label{eqar}
\end{equation}
We define the autoregressive state at step $t$ as $s_t=(x,y_{<t})$. Given a shared tokenizer, the student network (parameterized by $\theta$) and the frozen teacher network define the next-token predictive distributions $\pi_S(\cdot\mid s_t;\theta)$ and $\pi_T(\cdot\mid s_t)$, respectively.

In structured visual prediction, the generated tokens may encode diverse outputs such as text, spatial coordinates, or discrete actions. Because each committed token updates the subsequent state, an incorrect prediction corrupts not only the immediate output but also the future contexts upon which downstream predictions and supervision rely.

\subsection{On-Policy Distillation and State Visitation}

OPD constructs training trajectories by sampling directly from the student policy,
\begin{equation}
\hat y_t \sim \pi_S(\cdot\mid \hat s_t;\theta),
\end{equation}
and aligns the student with the teacher over these actively visited states:
\begin{equation}
\mathcal{L}_{\mathrm{OPD}}
=
\frac{1}{\sum_t m_t}
\sum_{t=1}^{L} m_t
D_{\mathrm{KL}}\left(
\pi_T(\cdot\mid \hat s_t)
\,\middle\|\,
\pi_S(\cdot\mid \hat s_t;\theta)
\right),
\label{eqopd}
\end{equation}
where $m_t$ denotes the loss mask. 

While this mitigates the distribution shift between training and free-running inference, unreliable student predictions, especially early in training, can severely degrade the quality of subsequent states.
To address this, we explicitly decouple the token-level supervision objective from the behavior policy $\mu$ used to construct the rollout:
\begin{equation}
y_t\sim\mu(\cdot\mid s_t),
\qquad
s_{t+1}=(x,y_{\le t}).
\label{eqmu}
\end{equation}
Standard OPD corresponds to $\mu=\pi_S$, which fully exposes the trajectory to student errors. Conversely, interleaved distillation permits teacher intervention during prefix construction. CA-OPD builds upon this generalized formulation by adaptively gating when a student proposal is retained. This dynamically balances the need for reliable, high-quality training trajectories with the necessity of gradually exposing the student to its own generated states.

\section{Confidence-Aware On-Policy Distillation}
\label{secmethod}

Building on Section~\ref{secproblem}, CA-OPD constructs the rollout prefix dynamically. At each decoding step, the teacher evaluates the student's proposed token. Highly supported proposals are retained and supervised via the teacher's predictive distribution, whereas unreliable ones are replaced by the teacher's prediction and supervised via direct cross-entropy. This single intervention decision elegantly aligns rollout control with adaptive token-level supervision.

\subsection{Confidence-Aware Rollout Control}
\label{secgate}

At state $s_t$, the student generates a candidate token $\hat y_t\sim\pi_S(\cdot\mid s_t;\theta)$. We measure the teacher's support for this exact proposal using the negative log-likelihood (NLL) defined as

\begin{equation}
\operatorname{NLL}_{T,t}
=
-\log \pi_T(\hat y_t\mid s_t)
\label{eqteachernll}
\end{equation}

A lower $\operatorname{NLL}_{T,t}$ indicates stronger teacher support. Given an NLL threshold $\tau_{\mathrm{nll},s}$ at optimization step $s$, we define a binary replacement indicator as

\begin{equation}
r_t
=
\mathbb I
\left[
\operatorname{NLL}_{T,t}
>
\tau_{\mathrm{nll},s}
\right]
\label{eqreplacementmask}
\end{equation}

Here, $r_t=1$ indicates the proposal is rejected, and $r_t=0$ means it is retained. Equivalently, this corresponds to a teacher probability floor $\alpha_s=\exp(-\tau_{\mathrm{nll},s})$. When a proposal is rejected, the teacher provides a deterministic correction $\tilde y_t=\arg\max_{v\in\mathcal V}\pi_T(v\mid s_t)$. The final token committed to the rollout is therefore

\begin{equation}
y_t
=
\begin{cases}
\tilde y_t, & r_t=1,\\[3pt]
\hat y_t, & r_t=0.
\end{cases}
\label{eqcommit}
\end{equation}

Generation subsequently continues from $s_{t+1}=(x,y_{\leq t})$. Because the prefix may contain both retained student tokens and teacher corrections, an intervention alters not only $y_t$ but also the autoregressive state conditioning all future predictions. These are discrete rollout operations, and no gradients are propagated through them.

\subsection{Intervention-Aligned Supervision}
\label{secloss}

The replacement indicator $r_t$ also dictates the token-level supervision applied at each position.

\paragraph{Supervision at replaced positions.}

For a replaced position ($r_t=1$), the committed token is the teacher's prediction $y_t=\tilde y_t$. The student is directly supervised using standard cross-entropy as

\begin{equation}
\mathcal L_t^{\mathrm R}
=
-\log \pi_S(y_t\mid s_t;\theta)
\label{eqreplacementloss}
\end{equation}

\paragraph{Supervision at retained positions.}

For a retained position ($r_t=0$), the student proposal remains. The student is then supervised by the teacher's predictive distribution via the forward KL divergence formulated as

\begin{equation}
\mathcal L_t^{\mathrm K}
=
D_{\mathrm{KL}}
\left(
\pi_T(\cdot\mid s_t)
\,\middle\|\,
\pi_S(\cdot\mid s_t;\theta)
\right).
\label{eqfullkl}
\end{equation}

To mitigate the computational cost of transferring the full vocabulary distribution, we distill only the teacher's top-$k$ predictions $\mathcal V_{T,t}^{(k)}=\operatorname{TopK}_{v}\pi_T(v\mid s_t)$. The truncated forward KL term is computed as

\begin{equation}
\mathcal L_t^{\mathrm{K\text{-}topk}}
=
\sum_{v\in\mathcal V_{T,t}^{(k)}}
\pi_T(v\mid s_t)
\left[
\log \pi_T(v\mid s_t)
-
\log \pi_S(v\mid s_t;\theta)
\right]
\label{eqtopkl}
\end{equation}

Because the retained probabilities are unnormalized, this truncated value can occasionally become negative. We therefore apply a zero lower bound where $\mathcal L_t^{\mathrm{K\text{-}topk}} \leftarrow \max(\mathcal L_t^{\mathrm{K\text{-}topk}},0)$.

\subsection{CA-OPD Objective}
\label{secunifiedloss}

By unifying rollout construction and supervision under the same replacement indicator, the token-level CA-OPD objective becomes

\begin{equation}
\mathcal L_t^{\mathrm{CA\text{-}OPD}
}
=
r_t\mathcal L_t^{\mathrm R}
+
(1-r_t)\mathcal L_t^{\mathrm{K\text{-}topk}}
\label{eqtokenloss}
\end{equation}

Let $m_t\in\{0,1\}$ denote the valid response token mask. The sequence-level objective is formulated as

\begin{equation}
\mathcal L_{\mathrm{CA\text{-}OPD}}
=
\frac{1}{\sum_{t=1}^{L}m_t}
\sum_{t=1}^{L}
m_t
\left[
r_t
\left(
-\log \pi_S(y_t\mid s_t;\theta)
\right)
+
(1-r_t)
\mathcal L_t^{\mathrm{K\text{-}topk}}
\right]
\label{eqcaopdloss}
\end{equation}

To provide further flexibility, these two supervision terms can be weighted independently as

\begin{equation}
\mathcal L_{\mathrm{CA\text{-}OPD}}
=
\frac{1}{\sum_{t=1}^{L}m_t}
\sum_{t=1}^{L}
m_t
\left[
\lambda_{\mathrm{CE}}
r_t\mathcal L_t^{\mathrm R}
+
\lambda_{\mathrm{KL}}
(1-r_t)\mathcal L_t^{\mathrm{K\text{-}topk}}
\right]
\label{eqweightedcaopd}
\end{equation}

\subsection{Progressive Transfer of Rollout Control}
\label{secschedule}

Early in training, aggressive teacher intervention prevents erratic student transitions from corrupting subsequent states. As the student matures, retaining more student-generated states becomes essential to preserve the on-policy nature of distillation. We therefore progressively relax the confidence requirement. 

Let $p_s=s/S$ denote the normalized training progress over $S$ total steps. We employ a cosine schedule to decay the intervention threshold according to

\begin{equation}
\tau_{\mathrm{nll},s}
=
\tau_{\mathrm{start}}
+
\frac{\tau_{\mathrm{end}}-\tau_{\mathrm{start}}}{2}
\left[
1-\cos(\pi p_s)
\right]
\label{eqnllschedule}
\end{equation}

where $\tau_{\mathrm{start}}<\tau_{\mathrm{end}}$. 
The lower initial threshold enforces a strict acceptance criterion. 
As training progresses, the threshold increases, permitting more student proposals to remain. 
Importantly, this schedule governs the confidence criterion rather than dictating a fixed replacement rate. 
As the student aligns more closely with the teacher, fewer positions trigger corrections, facilitating a natural transfer of rollout control.
The pseudocode for the experimental procedure is provided in Appendix~\ref{sectrainingprocedure}.

\section{Experiments}
\label{secexp}

\subsection{Experimental Setup}
\label{secsettings}

\paragraph{Models.}
We employ an SFT-initialized Qwen3.5-0.8B~\citep{qwen3.5} as the student. For supervision, we utilize two separate Qwen3.5-9B models, fine-tuned on their respective domains, to serve as the grounding and OCR teachers. During training, examples are dynamically routed to the corresponding domain teacher.

\paragraph{Baselines.}
We compare CA-OPD against three representative distillation strategies: \emph{Offline KD}, which trains the student on static teacher-generated trajectories~\citep{hinton2015distilling,agarwal2024policy}; \emph{OPD}, which performs standard on-policy distillation without teacher intervention~\citep{agarwal2024policy}; and \emph{SKD}~\citep{xu2025speculative}, which uses the teacher's relative ranking of each student proposal to determine whether it is retained in the rollout. We additionally evaluate a CA-OPD variant with a fixed acceptance threshold to isolate the contribution of our progressive schedule. The SFT-initialized student backbone and the teachers are reported as upper and lower reference points rather than controlled baselines.

\paragraph{Evaluation.}
We evaluate GUI grounding on ScreenSpot-v2~\citep{wu2025atlas} and ScreenSpot-Pro~\citep{li2025screenspot}, and OCR capabilities on OCRBench-v2~\citep{fu2026ocrbench}, CC-OCR~\citep{yang2025cc}, and OmniDocBench~\citep{ouyang2025omnidocbench}. 
Furthermore, we evaluate on RefCOCO/+/g~\citep{yu2016modeling,nagaraja2016modeling} and MMBench~\citep{liu2024mmbench} to measure general capability retention. 
Beyond standard benchmarking, our evaluation includes extensive controlled studies to isolate the impact of individual CA-OPD components. Detailed  protocols are provided in Appendix~\ref{appbenchmarks}. 

\paragraph{Implementation Details.}
To ensure fair comparison, all evaluated methods share the exact same student initialization, training data, and optimization budget. Unless otherwise stated, all reported results are averaged over three independent training runs. Detailed SFT and distillation data compositions, hyperparameters, decoding settings, and specific CA-OPD configurations are deferred to Appendix~\ref{appimplementation}.

\subsection{Main Results}
\label{secmain}

\begin{table*}[t]
\centering
\caption{
\textbf{Main results across target and retention benchmarks.}
All controlled distillation methods use the same student initialization,
training data, and optimization budget, and results are averaged over three
independent training runs.
For the teacher reference row, $^{\mathrm{G}}$ denotes the grounding teacher
and $^{\mathrm{O}}$ denotes the OCR teacher.
Best and second-best student results are shown in \textbf{bold} and
\underline{underline}, respectively.
}
\label{tabmain}

\begingroup
\renewcommand{\arraystretch}{1.08}
\setlength{\tabcolsep}{3.5pt}

\resizebox{\textwidth}{!}{%
\begin{tabular}{lcccccccc}
\toprule

\rowcolor{tablegray}
&
\multicolumn{2}{c}{GUI Grounding}
& \multicolumn{4}{c}{OCR}
& \multicolumn{2}{c}{Retention} \\
\noalign{\vskip -3pt}
\cmidrule(lr){2-3}
\cmidrule(lr){4-7}
\cmidrule(lr){8-9}
\noalign{\vskip -3pt}

\rowcolor{tablegray}
Method
Method
& SS-v2 & SS-Pro & OCRv2-EN & OCRv2-ZH & CC-OCR & OmniDoc
& RefCOCO
& MMBench \\
\midrule

\rowcolor{tablegray}
Teacher (domain-routed, 9B)
& 93.88$^{\mathrm{G}}$
& 62.92$^{\mathrm{G}}$
& 66.03$^{\mathrm{O}}$
& 63.87$^{\mathrm{O}}$
& 78.52$^{\mathrm{O}}$
& 81.18$^{\mathrm{O}}$
& 84.56$^{\mathrm{G}}$
& 90.60$^{\mathrm{G}}$ \\

\rowcolor{tablegray}
Qwen3.5-0.8B
& 79.33 & 36.42
& 45.75 & 43.45
& 62.19 & 62.28
& 78.51 & 75.21 \\

\rowcolor{tablegray}
Student (SFT-init)
& 80.46 & 38.19
& 50.98 & 51.31
& 66.40 & \underline{70.49}
& \textbf{82.00} & 79.16 \\

\midrule

\rowcolor{tablegray}
Offline KD~\citep{hinton2015distilling}
& \underline{87.96} & 44.23
& 51.56 & \underline{52.91}
& \underline{66.99} & 70.40
& 81.86 & 79.99 \\

\rowcolor{tablegray}
OPD~\citep{agarwal2024policy}
& 87.34 & 40.82
& 50.79 & 52.45
& 63.93 & 68.57
& 81.78 & \underline{80.00} \\

\rowcolor{tablegray}
SKD~\citep{xu2025speculative}
& 87.63 & 41.20
& 51.30 & 52.43
& 63.83 & 68.36
& 81.81 & 79.76 \\

\rowcolor{tablegray}
CA-OPD, fixed $\alpha=0.25$
& 87.81 & \underline{44.58}
& \underline{52.44} & 52.76
& 66.95 & \textbf{70.64}
& 81.82 & 79.73 \\

\rowcolor{caopdblue}
\textbf{CA-OPD, annealed $\alpha:0.50\!\rightarrow\!0.25$}
& \textbf{88.41} & \textbf{45.92}
& \textbf{52.47} & \textbf{53.22}
& \textbf{67.32} & 70.43
& \underline{81.90} & \textbf{80.16} \\

\bottomrule
\end{tabular}%
}

\endgroup
\end{table*}

Table~\ref{tabmain} summarizes the overall results. 
CA-OPD consistently outperforms standard OPD across all six target benchmarks spanning both GUI grounding and OCR. 
Crucially, the gains are substantial on complex tasks. 
For instance, ScreenSpot-Pro surges from $40.82$ to $45.92$ ($+5.10$ points), and CC-OCR increases from $63.93$ to $67.32$ ($+3.39$ points). 
Furthermore, CA-OPD surpasses both SKD and Offline KD across the board. 
This confirms that our improvements stem directly from the confidence-aware intervention mechanism, rather than the mere presence of on-policy supervision or teacher-generated trajectories.

Validating our dynamic intervention strategy, the annealed variant proves superior to a fixed confidence threshold. 
It improves five of the six target metrics, which supports the core hypothesis that progressively returning rollout control to the student as it matures is more effective than a static policy. 
Importantly, these target-task gains do not compromise generalization. 
When deliberately evaluated on held-out out-of-distribution (OOD) tasks such as RefCOCO and MMBench, CA-OPD maintains performance on par with the SFT initialization and competing baselines, successfully avoiding catastrophic forgetting.
% Detailed results are shown in Appendix \ref{secres}.
Detailed results and an additional evaluation with a
Qwen3-VL teacher--student pair are reported in Appendices~\ref{secres} and~\ref{appqwen3vl}, respectively.
% To assess robustness beyond the main Qwen3.5 setting, we evaluate CA-OPD with an additional Qwen3-VL teacher--student pair; full results are reported in Appendix~\ref{appqwen3vl}.

\subsection{Ablation study}
\label{seccomponents}

\begin{table}[t]
\centering
\caption{
\textbf{Ablation study of CA-OPD.}
We separately examine selective prefix repair and the remaining design choices for rollout control and supervision.
\textit{WB} denotes the realized fraction of rollout tokens replaced by teacher predictions. Random intervention uses a comparable writeback budget.
}
\label{tabcoreablation}

\begingroup
\fontsize{7.5pt}{8.5pt}\selectfont
\renewcommand{\arraystretch}{1.08}

% =========================================================
% Selective intervention and prefix repair
% =========================================================
\textbf{Selective Intervention and Prefix Repair}

\vspace{1.5pt}

\setlength{\tabcolsep}{4.2pt}

\begin{tabular}{lccccc}
\toprule
Variant
& Placement
& \makecell{Prefix\\Repair}
& SS-Pro
& OCRv2-EN
& WB (\%) \\
\midrule

OPD
& None
& No
& 40.82
& 50.79
& 0.00 \\

Random intervention
& Random
& Yes
& 40.20
& 50.62
& 3.87 \\

No prefix writeback
& Confidence
& No
& 40.92
& 51.36
& 0.00 \\

\rowcolor{caopdblue}
\textbf{CA-OPD}
& \textbf{Confidence}
& \textbf{Yes}
& \textbf{45.92}
& \textbf{52.47}
& 5.40 \\

\bottomrule
\end{tabular}

\vspace{5pt}

% =========================================================
% Rollout control and supervision
% =========================================================
\textbf{Rollout Control and Supervision}

\vspace{1.5pt}

\setlength{\tabcolsep}{3.4pt}

\begin{tabular}{lcccccc}
\toprule
Variant
& Schedule
& Correction
& \makecell{Loss at\\Replaced}
& SS-Pro
& OCRv2-EN
& WB (\%) \\
\midrule

Fixed threshold
& Fixed
& Argmax
& CE
& 44.58
& 52.44
& 4.90 \\

Reversed schedule
& Reversed
& Argmax
& CE
& 44.36
& 51.69
& 4.90 \\

Sampled writeback
& Annealed
& Sample
& CE
& 43.61
& 51.69
& 5.60 \\

KL at replaced positions
& Annealed
& Argmax
& KL
& 45.26
& 51.75
& 5.40 \\

\rowcolor{caopdblue}
\textbf{CA-OPD}
& \textbf{Annealed}
& \textbf{Argmax}
& \textbf{CE}
& \textbf{45.92}
& \textbf{52.47}
& 5.40 \\

\bottomrule
\end{tabular}
\endgroup
\end{table}

We ablate CA-OPD in two stages. 
We first examine whether the performance gains depend on both explicit prefix repair and intervention placement guided by confidence. 
We then retain these two core components to isolate the individual effects of rollout scheduling, teacher correction policies, and supervision aligned with intervention decisions. 
All variants maintain identical student initialization, training data, and optimization budgets.

\paragraph{Selective intervention and prefix repair are both essential.}
The first block of Table~\ref{tabcoreablation} separates the effect of where the teacher intervenes from whether the correction is actually written back into the autoregressive prefix. 
Randomly inserting teacher corrections at a comparable writeback rate performs similarly to standard OPD. 
This indicates that merely increasing teacher intervention is insufficient. Conversely, applying confidence gating without prefix writeback also yields performance close to standard OPD, despite preserving the intervention decision and the associated supervision. 
Only the combination of selective placement and actual prefix repair produces the full improvement. 
This result demonstrates that CA-OPD benefits heavily from selectively correcting unreliable student transitions and using those corrections to improve the subsequent states visited during training.

\paragraph{Rollout control and supervision provide complementary gains.}
The second block evaluates the remaining design choices while retaining selective intervention and prefix repair. 
Replacing the annealed confidence schedule with a fixed threshold degrades performance, and reversing the schedule causes an even larger drop. 
This finding strongly supports the progressive transfer of rollout control from the teacher to the student. 
Furthermore, using sampled teacher corrections instead of deterministic argmax writeback reduces performance, suggesting that reliable prefix repair requires a stable correction policy. 
Finally, replacing direct corrective supervision at replaced positions with KL distillation underperforms the full objective. 
% Together, these ablations prove that progressive scheduling, deterministic writeback, and adaptive supervision each contribute distinct gains on top of selective prefix repair.

\subsection{Analysis of the Gating Mechanism}
\label{secgateanalysis}
\label{secbudget}
\label{secriskvaliditymain}
\label{secdynamicsmain}

We examine whether CA-OPD's gains arise simply from more frequent
interventions, whether absolute teacher support identifies positions missed
by rank-based gating, and whether these interventions improve the resulting
trajectories. Together, these analyses isolate whether the benefit comes from
the intervention budget itself or from selectively correcting particular
rollout states.

\begin{figure*}[t]
\centering
\includegraphics[
    width=0.99\textwidth,
    trim=0 0 0 0,
    clip
]{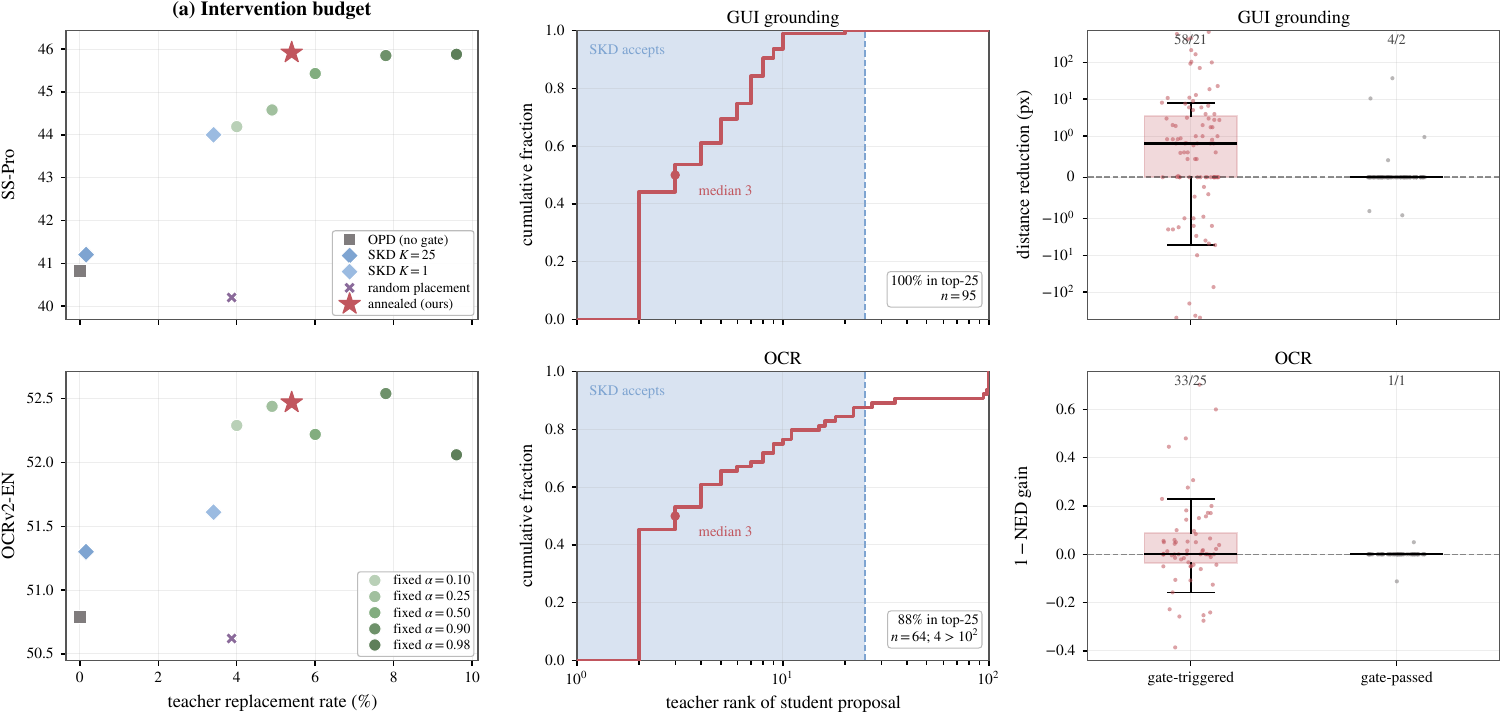}
\vspace{-3pt}
\caption{
\textbf{Analysis of the CA-OPD gating mechanism.}
\textbf{(a)} Increasing the intervention budget alone does not recover
CA-OPD's gains.
\textbf{(b)} Most CA-OPD-triggered proposals remain within the teacher's
top-$25$ despite insufficient absolute support.
\textbf{(c)} Teacher replacement benefits gate-triggered positions, with
mostly unchanged gate-passed outcomes; annotations show better/worse
counts, excluding ties.
}
\label{figgatinganalysis}
\vspace{-6pt}
\end{figure*}

\paragraph{Intervention frequency alone does not explain the gains.}
As shown in Figure~\ref{figgatinganalysis}(a), increasing the SKD replacement
rate from $0.16\%$ to $3.41\%$ improves SS-Pro from $41.20$ to $44.00$,
but still trails CA-OPD by $1.92$ points on SS-Pro and $0.86$ points on
OCRv2-EN.
CA-OPD performance is also non-monotonic in replacement rate, while random
interventions at a comparable budget yield only marginal gains
(Table~\ref{tabcoreablation}).
Thus, intervention placement matters beyond frequency alone.
In particular, simply allocating more teacher tokens does not reproduce the
benefit of selectively intervening at low-support states.
The full budget sweep is provided in Appendix~\ref{appbudgetsweep}.

\paragraph{Absolute support identifies positions missed by rank.}
Figure~\ref{figgatinganalysis}(b) shows that CA-OPD-triggered proposals have
a median teacher rank of $3$, with $100\%$ of grounding and $88\%$ of OCR
proposals remaining within the teacher's top-$25$.
Thus, rank-based gating would retain most proposals rejected by CA-OPD due
to insufficient absolute teacher support.

\paragraph{Gate-triggered positions are consequential.}
Figure~\ref{figgatinganalysis}(c) compares paired student continuations
from the same prefix, retaining the proposed token or replacing it with
the teacher argmax.
Replacement improves $58$ of $95$ triggered grounding cases and yields
a positive mean effect on OCR, while gate-passed scores mostly remain
unchanged.
Complete counts, including unchanged outcomes, are reported in
Appendix Table~\ref{tabcounterfactualcounts}.

Finally, teacher intervention frequency naturally decreases during training,
keeping rollouts predominantly student-driven.
This decay indicates that the method does not rely on persistently high levels
of teacher correction as optimization proceeds.
Detailed training dynamics are provided in
Appendix~\ref{appgatingdynamics}.

\subsection{Stratified Performance Analysis}
\label{secstratifiedmain}

We examine whether the gains of CA-OPD are concentrated in particular
output-length, initial-competence, or teacher-rejection regimes.

\begin{table}[t]
\centering
\caption{
\textbf{CA-OPD gains across different data regimes.}
Each cell reports the improvement of annealed CA-OPD over OPD.
The gains remain positive across all strata rather than being confined to a
particular output length, initial competence level, or teacher-rejection density.
}
\label{tabstratifiedmain}

\begingroup
\fontsize{7.5pt}{8.5pt}\selectfont
\renewcommand{\arraystretch}{1.12}
\setlength{\tabcolsep}{5.5pt}

\begin{tabular}{lccc}
\toprule

\rowcolor{tablegray}
\textbf{Stratification}
& \textbf{Low / Short}
& \textbf{Medium}
& \textbf{High / Long} \\

\midrule

\textbf{Output length}
& \makecell[c]{Short\\\textbf{$+1.50$}}
& \makecell[c]{Medium\\\textbf{$+0.61$}}
& \makecell[c]{Long\\\textbf{$+1.16$}} \\

\addlinespace[3pt]

\textbf{Initial competence}
& \makecell[c]{All wrong\\\textbf{$+3.90$}}
& \makecell[c]{Mixed\\\textbf{$+10.60$}}
& \makecell[c]{All correct\\\textbf{$+3.65$}} \\

\addlinespace[3pt]

\textbf{Teacher-rejection density}
& \makecell[c]{Low $\rho$\\\textbf{$+4.95$}}
& \makecell[c]{Medium $\rho$\\\textbf{$+5.31$}}
& \makecell[c]{High $\rho$\\\textbf{$+5.05$}} \\

\bottomrule
\end{tabular}

\endgroup
\end{table}

As shown in Table~\ref{tabstratifiedmain}, CA-OPD improves over OPD in
every stratum, indicating that its gains are not confined to long outputs,
low-competence examples, or states with dense teacher rejection.
In particular, the nearly constant margin across rejection-density terciles
suggests that the benefit extends beyond examples requiring frequent
intervention.
Although the mixed-competence group exhibits the largest gain, we treat this
as descriptive rather than evidence of a specific intermediate-difficulty
effect.
CA-OPD also better preserves behavior already acquired during SFT.
On examples consistently solved by the initialization, OPD decreases
performance from $96.03$ to $90.18$, whereas CA-OPD retains $93.83$
(Appendix Table~\ref{tabinitcompetencefull}).
Detailed scores and stratification protocols are provided in
Appendix~\ref{appstratified} and Appendix~\ref{appstratprotocol}.

% \subsection{Discussion}
% \label{secdiscussionmain}

% Overall, the experiments show that CA-OPD benefits from selectively placing a small amount of teacher intervention rather than simply increasing the intervention budget.
% Its advantage comes from coupling rollout repair with intervention-aligned supervision: rejected student proposals are corrected with teacher tokens and trained using cross-entropy, while retained proposals preserve distributional distillation.
% The gate in fact separates two sharply different supervision regimes: at retained positions the teacher distribution is nearly deterministic (only $4\%$ of positions have entropy above $0.5$), whereas at replaced positions it is diffuse (mean entropy $1.0$--$1.3$) yet still concentrates on a decisive correction (median argmax probability $0.55$--$0.68$), which is where a hard target is most useful.
% The strict-to-relaxed schedule further provides stronger teacher control early in training and progressively returns control to the student, while the gains observed across output-length, initial-competence, and teacher-endorsement strata suggest that CA-OPD improves autoregressive training states broadly rather than only mitigating long-horizon error propagation.
% These results support viewing CA-OPD as a selective state-control mechanism that preserves the on-policy nature of OPD while reducing the effect of poorly supported student transitions.

\section{Conclusion}
\label{secconclusion}

We introduced CA-OPD, a confidence-aware on-policy distillation framework
for structured visual prediction that uses teacher confidence to selectively
repair student-generated trajectories and couples rollout intervention with
token-level supervision.
Across GUI grounding and OCR, CA-OPD consistently improves over standard OPD
and competing distillation methods while preserving held-out capabilities.
Controlled analyses further show that its gains depend on selective
intervention placement, progressive rollout control, and
intervention-aligned supervision, highlighting the importance of jointly
considering trajectory construction and knowledge transfer in on-policy
distillation.

\clearpage

\subsection*{AI use statement}

In this work, we used generative AI tools as coding assistants for parts of
the experimental implementation. AI-generated code was manually reviewed,
modified when necessary, and tested by the authors to ensure correctness and
reproducibility.
We did not use generative AI tools for developing the research ideas,
formulating the problem, designing the proposed method, deriving the
algorithm, interpreting experimental results, or making scientific claims.
We also did not use generative AI tools to generate experimental conclusions
or replace the authors' research judgment.
Additionally, generative AI tools were used for limited language refinement
during manuscript preparation. All AI-assisted content was carefully reviewed
and revised by the authors. We take full responsibility for the final content
of this work, including all text, claims, experimental results, and artifacts
produced with the aid of generative AI.

\bibliography{iclr2025_conference}

@inproceedings{kim2022ocr,
  title={Ocr-free document understanding transformer},
  author={Kim, Geewook and Hong, Teakgyu and Yim, Moonbin and Nam, JeongYeon and Park, Jinyoung and Yim, Jinyeong and Hwang, Wonseok and Yun, Sangdoo and Han, Dongyoon and Park, Seunghyun},
  booktitle={European Conference on Computer Vision},
  pages={498--517},
  year={2022},
  organization={Springer}
}

@article{yuan2026vision,
  title={Vision-opd: Learning to see fine details for multimodal llms via on-policy self-distillation},
  author={Yuan, Qianhao and Lou, Jie and Yu, Xing and Lin, Hongyu and Sun, Le and Han, Xianpei and Lu, Yaojie},
  journal={arXiv preprint arXiv:2605.18740},
  year={2026}
}

@inproceedings{wu2025atlas,
  title={OS-ATLAS: Foundation action model for generalist GUI agents},
  author={Wu, Zhiyong and Wu, Zhenyu and Xu, Fangzhi and Wang, Yian and Sun, Qiushi and Jia, Chengyou and Cheng, Kanzhi and Ding, Zichen and Chen, Liheng and Liang, Paul Pu and others},
  booktitle={International Conference on Learning Representations},
  volume={2025},
  pages={5090--5108},
  year={2025}
}

@inproceedings{li2025screenspot,
  title={Screenspot-pro: Gui grounding for professional high-resolution computer use},
  author={Li, Kaixin and Meng, Ziyang and Lin, Hongzhan and Luo, Ziyang and Tian, Yuchen and Ma, Jing and Huang, Zhiyong and Chua, Tat-Seng},
  booktitle={Proceedings of the 33rd ACM International Conference on Multimedia},
  pages={8778--8786},
  year={2025}
}

@article{fu2026ocrbench,
  title={Ocrbench v2: An improved benchmark for evaluating large multimodal models on visual text localization and reasoning},
  author={Fu, Ling and Kuang, Zhebin and Song, Jiajun and Huang, Mingxin and Yang, Biao and Li, Yuzhe and Zhu, Linghao and Luo, Qidi and Wang, Xinyu and Lu, Hao and others},
  journal={Advances in Neural Information Processing Systems},
  volume={38},
  year={2026}
}

@inproceedings{yang2025cc,
  title={Cc-ocr: A comprehensive and challenging ocr benchmark for evaluating large multimodal models in literacy},
  author={Yang, Zhibo and Tang, Jun and Li, Zhaohai and Wang, Pengfei and Wan, Jianqiang and Zhong, Humen and Liu, Xuejing and Yang, Mingkun and Wang, Peng and Bai, Shuai and others},
  booktitle={2025 IEEE/CVF International Conference on Computer Vision (ICCV)},
  pages={21744--21754},
  year={2025},
  organization={IEEE}
}

@inproceedings{ouyang2025omnidocbench,
  title={Omnidocbench: Benchmarking diverse pdf document parsing with comprehensive annotations},
  author={Ouyang, Linke and Qu, Yuan and Zhou, Hongbin and Zhu, Jiawei and Zhang, Rui and Lin, Qunshu and Wang, Bin and Zhao, Zhiyuan and Jiang, Man and Zhao, Xiaomeng and others},
  booktitle={2025 IEEE/CVF Conference on Computer Vision and Pattern Recognition (CVPR)},
  pages={24838--24848},
  year={2025},
  organization={IEEE}
}

@inproceedings{yu2016modeling,
  title={Modeling context in referring expressions},
  author={Yu, Licheng and Poirson, Patrick and Yang, Shan and Berg, Alexander C and Berg, Tamara L},
  booktitle={European conference on computer vision},
  pages={69--85},
  year={2016},
  organization={Springer}
}

@inproceedings{nagaraja2016modeling,
  title={Modeling context between objects for referring expression understanding},
  author={Nagaraja, Varun K and Morariu, Vlad I and Davis, Larry S},
  booktitle={European conference on computer vision},
  pages={792--807},
  year={2016},
  organization={Springer}
}

@inproceedings{liu2024mmbench,
  title={Mmbench: Is your multi-modal model an all-around player?},
  author={Liu, Yuan and Duan, Haodong and Zhang, Yuanhan and Li, Bo and Zhang, Songyang and Zhao, Wangbo and Yuan, Yike and Wang, Jiaqi and He, Conghui and Liu, Ziwei and others},
  booktitle={European conference on computer vision},
  pages={216--233},
  year={2024},
  organization={Springer}
}

@misc{qwen3.5,
    title  = {{Qwen3.5}: Towards Native Multimodal Agents},
    author = {{Qwen Team}},
    year   = {2026},
    month  = {February},
    url    = {https://qwen.ai/blog?id=qwen3.5}
}

@article{ranzato2015sequence,
  title={Sequence level training with recurrent neural networks},
  author={Ranzato, Marc'Aurelio and Chopra, Sumit and Auli, Michael and Zaremba, Wojciech},
  journal={arXiv preprint arXiv:1511.06732},
  year={2015}
}

@article{wang2026stepx,
  title={StepX-Edge: An On-Device UI Vision-Language Model via Architecture-Training-Deployment Co-Design},
  author={Wang, Yin and Hu, Haotian and Han, Jineng and Qiu, Wentao and Ge, Zhenhua and Tang, Liujian and Wang, Fanyi},
  journal={arXiv preprint arXiv:2607.22708},
  year={2026}
}

@article{chen2021pix2seq,
  title={Pix2seq: A language modeling framework for object detection},
  author={Chen, Ting and Saxena, Saurabh and Li, Lala and Fleet, David J and Hinton, Geoffrey},
  journal={arXiv preprint arXiv:2109.10852},
  year={2021}
}

@article{kim2021donut,
  title={Donut: document understanding transformer without OCR. CoRR abs/2111.15664 (2021)},
  author={Kim, Geewook and Hong, Teakgyu and Yim, Moonbin and Park, Jinyoung and Yim, Jinyeong and Hwang, Wonseok and Yun, Sangdoo and Han, Dongyoon and Park, Seunghyun},
  journal={arXiv preprint arXiv:2111.15664},
  year={2021}
}

@article{lu2022unified,
  title={Unified-io: A unified model for vision, language, and multi-modal tasks},
  author={Lu, Jiasen and Clark, Christopher and Zellers, Rowan and Mottaghi, Roozbeh and Kembhavi, Aniruddha},
  journal={arXiv preprint arXiv:2206.08916},
  year={2022}
}

@inproceedings{lu2024unified,
  title={Unified-io 2: Scaling autoregressive multimodal models with vision, language, audio, and action},
  author={Lu, Jiasen and Clark, Christopher and Lee, Sangho and Zhang, Zichen and Khosla, Savya and Marten, Ryan and Hoiem, Derek and Kembhavi, Aniruddha},
  booktitle={2024 IEEE/CVF Conference on Computer Vision and Pattern Recognition (CVPR)},
  pages={26429--26445},
  year={2024},
  organization={IEEE}
}

@article{peng2023kosmos,
  title={Kosmos-2: Grounding multimodal large language models to the world},
  author={Peng, Zhiliang and Wang, Wenhui and Dong, Li and Hao, Yaru and Huang, Shaohan and Ma, Shuming and Wei, Furu},
  journal={arXiv preprint arXiv:2306.14824},
  year={2023}
}

@article{chen2023shikra,
  title   = {{Shikra}: Unleashing Multimodal {LLM}'s Referential Dialogue Magic},
  author  = {Chen, Keqin and Zhang, Zhao and Zeng, Weili and Zhang, Richong and Zhu, Feng and Zhao, Rui},
  journal = {arXiv preprint arXiv:2306.15195},
  year    = {2023}
}

@inproceedings{you2024ferret,
  title={Ferret: Refer and ground anything anywhere at any granularity},
  author={You, Haoxuan and Zhang, Haotian and Gan, Zhe and Du, Xianzhi and Zhang, Bowen and Wang, Zirui and Cao, Liangliang and Chang, Shih-Fu and Yang, Yinfei},
  booktitle={International Conference on Learning Representations},
  volume={2024},
  pages={57153--57180},
  year={2024}
}

@inproceedings{cheng2024seeclick,
  title     = {{SeeClick}: Harnessing {GUI} Grounding for Advanced Visual {GUI} Agents},
  author    = {Cheng, Kanzhi and Sun, Qiushi and Chu, Yougang and Xu, Fangzhi and Li, Yantao and Zhang, Jianbing and Wu, Zhiyong},
  booktitle = {Proceedings of the 62nd Annual Meeting of the Association for Computational Linguistics},
  pages     = {9313--9332},
  year      = {2024},
  doi       = {10.18653/v1/2024.acl-long.505}
}

@inproceedings{lin2025showui,
  title     = {{ShowUI}: One Vision-Language-Action Model for {GUI} Visual Agent},
  author    = {Lin, Kevin Qinghong and Li, Linjie and Gao, Difei and Yang, Zhengyuan and Wu, Shiwei and Bai, Zechen and Lei, Stan Weixian and Wang, Lijuan and Shou, Mike Zheng},
  booktitle = {Proceedings of the IEEE/CVF Conference on Computer Vision and Pattern Recognition},
  pages     = {19498--19508},
  year      = {2025},
  doi       = {10.1109/CVPR52734.2025.01816}
}

@article{hinton2015distilling,
  title={Distilling the Knowledge in a Neural Network},
  author={Hinton, Geoffrey and Vinyals, Oriol and Dean, Jeff},
  journal={arXiv preprint arXiv:1503.02531},
  year={2015}
}

@inproceedings{kim2016sequence,
  title={Sequence-level knowledge distillation},
  author={Kim, Yoon and Rush, Alexander M},
  booktitle={Proceedings of the 2016 conference on empirical methods in natural language processing},
  pages={1317--1327},
  year={2016}
}

@inproceedings{hsieh2023distilling,
  title     = {Distilling Step-by-Step! Outperforming Larger Language Models with Less Training Data and Smaller Model Sizes},
  author    = {Hsieh, Cheng-Yu and Li, Chun-Liang and Yeh, Chih-Kuan and Nakhost, Hootan and Fujii, Yasuhisa and Ratner, Alex and Krishna, Ranjay and Lee, Chen-Yu and Pfister, Tomas},
  booktitle = {Findings of the Association for Computational Linguistics: ACL 2023},
  pages     = {8003--8017},
  year      = {2023},
  doi       = {10.18653/v1/2023.findings-acl.507}
}

@inproceedings{gu2024minillm,
  title={Minillm: Knowledge distillation of large language models},
  author={Gu, Yuxian and Dong, Li and Wei, Furu and Huang, Minlie},
  booktitle={International Conference on Learning Representations},
  volume={2024},
  pages={32694--32717},
  year={2024}
}

@inproceedings{cai2025llavakd,
  title     = {{LLaVA-KD}: A Framework of Distilling Multimodal Large Language Models},
  author    = {Cai, Yuxuan and Zhang, Jiangning and He, Haoyang and He, Xinwei and Tong, Ao and Gan, Zhenye and Wang, Chengjie and Xue, Zhucun and Liu, Yong and Bai, Xiang},
  booktitle = {Proceedings of the IEEE/CVF International Conference on Computer Vision},
  pages     = {239--249},
  year      = {2025},
  doi       = {10.1109/ICCV51701.2025.00030}
}

@article{bengio2015scheduled,
  title={Scheduled sampling for sequence prediction with recurrent neural networks},
  author={Bengio, Samy and Vinyals, Oriol and Jaitly, Navdeep and Shazeer, Noam},
  journal={Advances in neural information processing systems},
  volume={28},
  year={2015}
}

@inproceedings{ross2011reduction,
  title={A reduction of imitation learning and structured prediction to no-regret online learning},
  author={Ross, St{\'e}phane and Gordon, Geoffrey and Bagnell, Drew},
  booktitle={Proceedings of the fourteenth international conference on artificial intelligence and statistics},
  pages={627--635},
  year={2011},
  organization={JMLR Workshop and Conference Proceedings}
}

@inproceedings{agarwal2024policy,
  title={On-policy distillation of language models: Learning from self-generated mistakes},
  author={Agarwal, Rishabh and Vieillard, Nino and Zhou, Yongchao and Stanczyk, Piotr and Ramos Garea, Sabela and Geist, Matthieu and Bachem, Olivier},
  booktitle={International Conference on Learning Representations},
  volume={2024},
  pages={21246--21263},
  year={2024}
}

@inproceedings{xu2025speculative,
  title={Speculative knowledge distillation: Bridging the teacher-student gap through interleaved sampling},
  author={Xu, Wenda and Han, Rujun and Wang, Zifeng and Le, Long and Madeka, Dhruv and Li, Lei and Wang, William and Agarwal, Rishabh and Lee, Chen-Yu and Pfister, Tomas},
  booktitle={International Conference on Learning Representations},
  volume={2025},
  pages={64616--64646},
  year={2025}
}
\bibliographystyle{iclr2025_conference}

\clearpage
\appendix

\section{Additional Method and Training Details}
\label{sectrainingprocedure}

Section~\ref{secmethod} presents the core CA-OPD formulation. This section provides the surrounding objectives, the exact warm-up adjustment used by the annealing schedule, and a single consolidated training algorithm.

\subsection{Objectives and Rollout Context}
\label{appobjectives}

Supervised fine-tuning conditions on an annotated prefix $y^{\star}_{<t}$ and minimizes
\begin{equation}
\mathcal{L}_{\mathrm{SFT}}
=
-\frac{1}{\sum_t m_t}
\sum_{t=1}^{L}m_t\log\pi_S(y^{\star}_t\mid x,y^{\star}_{<t};\theta).
\label{eqsft}
\end{equation}
Offline distillation instead uses teacher-generated prefixes $y^{\mathrm{off}}_{<t}$ and matches the teacher distribution:
\begin{equation}
\mathcal{L}_{\mathrm{off}}
=
\frac{1}{\sum_t m_t}
\sum_{t=1}^{L}m_t
D_{\mathrm{KL}}\left(
\pi_T(\cdot\mid x,y^{\mathrm{off}}_{<t})
\,\middle\|\,
\pi_S(\cdot\mid x,y^{\mathrm{off}}_{<t};\theta)
\right).
\label{eqoffline}
\end{equation}
Thus, SFT and offline KD replay fixed prefixes, standard OPD delegates every transition to the student, and interleaved methods operate between these two endpoints. 
CA-OPD further couples this mixed state-visitation policy to the supervision rule: teacher-written positions receive cross-entropy, whereas retained student positions receive distributional distillation.

\subsection{Warm-Up-Adjusted Annealing Schedule}
\label{appschedule}

The threshold schedule in Equation~\ref{eqnllschedule} is evaluated using a warm-up-adjusted progress variable. For warm-up fraction $\omega$ and total
optimizer steps $S$,
\begin{equation}
p_s=
\begin{cases}
0, & s/S\le\omega,\\[3pt]
\dfrac{s/S-\omega}{1-\omega}, & s/S>\omega.
\end{cases}
\label{eqwarmup}
\end{equation}
The acceptance criterion is therefore held at its strictest setting during the first fraction $\omega$ of training, after which cosine relaxation begins.
The main configuration uses $\omega=0.1$, $\tau_{\mathrm{start}}=0.693$, and $\tau_{\mathrm{end}}=1.386$, corresponding to a teacher-probability floor
annealed from $\alpha=0.50$ to $\alpha=0.25$ over $471$ optimizer steps. 
This warm-up concerns rollout control only; the optimizer learning-rate schedule has no warm-up.

\subsection{Training Algorithm and Chunked Rollouts}
\label{appalgorithm}

Algorithm~\ref{algcaopd} gives the token-level semantics of rollout construction and intervention-aligned supervision.
The algorithm is written token by token for clarity. 
In the implementation, the student proposes up to $32$ tokens per chunk and a teacher query scores the proposed tokens under the shared prefix. 
If a replacement is triggered, generation restarts from the first replaced position, so every later chunk is conditioned on the updated mixed prefix. 
These chunked operations preserve the same sequential semantics while reducing query overhead. 
Their measured cost is reported in Appendix~\ref{appcompute}.

\begin{algorithm}[htbp]
\caption{Confidence-Aware On-Policy Distillation for one training example}
\label{algcaopd}
\begin{algorithmic}[1]
\Require input $x$, student $\pi_S$, frozen teacher $\pi_T$
\Require optimizer step $s$, total steps $S$, warm-up fraction $\omega$
\Require $\tau_{\mathrm{start}}$, $\tau_{\mathrm{end}}$, loss truncation $k$
\State $y_{<1}\gets\varnothing$
\State compute $p_s$ using Equation~\ref{eqwarmup}
\State compute $\tau_{\mathrm{nll},s}$ using Equation~\ref{eqnllschedule}
\For{$t=1$ \textbf{to} $L$}
    \State $s_t\gets(x,y_{<t})$
    \State sample $\hat y_t\sim\pi_S(\cdot\mid s_t;\theta)$
    \State evaluate $\pi_T(\cdot\mid s_t)$ and
    $\operatorname{NLL}_{T,t}\gets-\log\pi_T(\hat y_t\mid s_t)$
    \State $r_t\gets\mathbb I[\operatorname{NLL}_{T,t}>\tau_{\mathrm{nll},s}]$
    \If{$r_t=1$}
        \State $y_t\gets\arg\max_{v\in\mathcal V}\pi_T(v\mid s_t)$
        \State $\mathcal L_t\gets-\log\pi_S(y_t\mid s_t;\theta)$
    \Else
        \State $y_t\gets\hat y_t$
        \State compute $\mathcal L_t^{\mathrm{K\text{-}topk}}$ using
        Equation~\ref{eqtopkl} and clamp it below at zero
        \State $\mathcal L_t\gets\mathcal L_t^{\mathrm{K\text{-}topk}}$
    \EndIf
    \State $y_{<t+1}\gets(y_{<t},y_t)$
\EndFor
\State \Return $\left(\sum_{t=1}^{L}m_t\mathcal L_t\right)/
\left(\sum_{t=1}^{L}m_t\right)$
\end{algorithmic}
\end{algorithm}

\section{Implementation Details}
\label{appimplementation}

All controlled methods in Table~\ref{tabmain} use the same student initialization, training data, batch size, and optimizer-step budget. 
Table~\ref{tabhyper} collects the resolved configuration of the main annealed CA-OPD run and makes explicit the distinction between optimizer scheduling and gate scheduling.

\begin{table}[htbp]
\centering
\caption{
\textbf{Training and implementation configuration.}
The optimizer settings are shared by every controlled method in Table~\ref{tabmain}; gate-specific settings apply to CA-OPD variants.
}
\label{tabhyper}

\begingroup
% \apptablesetup

\begin{tabularx}{\linewidth}{@{}lX@{}}
\toprule

\rowcolor{tablegray}
Setting & Value \\
\midrule

% =========================================================
% Models and data
% =========================================================
\rowcolor{tablegray}
\multicolumn{2}{c}{\emph{Models and data}} \\

Student & SFT-initialized Qwen3.5-0.8B \\
Grounding teacher & Grounding-fine-tuned Qwen3.5-9B \\
OCR teacher & OCR-fine-tuned Qwen3.5-9B \\
SFT data & 40K OCR examples and 6K RefCOCO examples \\
Distillation data & 15,100 examples: 11,184 grounding and 3,916 OCR \\
Independent runs & 3 for every controlled method in Table~\ref{tabmain} \\

\addlinespace[2pt]

% =========================================================
% Optimization
% =========================================================
\rowcolor{tablegray}
\multicolumn{2}{c}{\emph{Optimization}} \\

Batch size & 32 \\
Optimizer steps & 471 \\
Learning rate & $1\times10^{-6}$ \\
Optimizer & AdamW, $\beta=(0.9,0.999)$ \\
Weight decay & 0.01 \\
Gradient clipping & 1.0 \\
Learning-rate schedule & Constant, with no optimizer warm-up \\
Precision & bfloat16 mixed precision with fp32 master weights \\

\addlinespace[2pt]

% =========================================================
% Rollout and supervision
% =========================================================
\rowcolor{tablegray}
\multicolumn{2}{c}{\emph{Rollout and supervision}} \\

Student rollout temperature & 1.0 \\
Teacher correction and decoding & Greedy teacher argmax \\
Gate schedule & Cosine $\alpha:0.50\!\rightarrow\!0.25$ after a $0.1$ hold fraction \\
Teacher top $k$ for gate scoring & 32 \\
Teacher top $k$ for retained-position loss & 64 \\
Maximum prompt length & 8,192 tokens \\
Maximum response length & 1,536 tokens \\

\addlinespace[2pt]

% =========================================================
% Systems
% =========================================================
\rowcolor{tablegray}
\multicolumn{2}{c}{\emph{Systems}} \\

Parallelism & FSDP, 4 GPUs for the student \\
Hardware & $8\times$ NVIDIA A800: 4 student GPUs and 4 teacher replicas \\

\bottomrule
\end{tabularx}

\endgroup
\end{table}

\subsection{Top-$k$ Teacher Transfer}
\label{apptopk}

The two teacher truncation settings in Table~\ref{tabhyper} serve different
purposes. During rollout construction, each gate query returns the teacher
log-probabilities of the proposed tokens in the current chunk under the
top-$32$ gate-scoring setting; the gate needs the probability of the exact
student proposal and, on rejection, the teacher argmax. At retained positions,
the distillation objective uses a forward KL restricted to the teacher's
top-$64$ tokens. The retained probabilities are not renormalized, the resulting
per-token value is clamped below at zero, and the retained teacher mass has a
training median above $0.999$, making the truncation immaterial in practice.

\subsection{Training Dataset Sources}
\label{appdatasets}

We use subsets of the publicly available datasets summarized in
Table~\ref{tabdatasets} for SFT initialization and distillation training.
Each dataset is processed and filtered according to the requirements of the
corresponding task domain. 

\begin{table}[t]
\centering
\footnotesize
\caption{
\textbf{Public datasets used in training.}
We use subsets of these datasets for SFT initialization and distillation
training.
}
\label{tabdatasets}

\renewcommand{\arraystretch}{1.05}

\begin{tabular}{lll}
\toprule
Dataset & Domain & Link \\
\midrule

\multicolumn{3}{c}{\textbf{GUI Grounding}} \\
\midrule

OS-Atlas
& Computer interaction
& \href{https://huggingface.co/datasets/OS-Copilot/OS-Atlas-data}{Hugging Face}
\\

RefCOCO
& Referring expression grounding
& \href{https://huggingface.co/datasets/lmms-lab-encoder/RefCOCO}{Hugging Face}
\\

\midrule

\multicolumn{3}{c}{\textbf{OCR and Document Understanding}} \\
\midrule

TextVQA
& Scene text understanding
& \href{https://huggingface.co/datasets/facebook/textvqa}{Hugging Face}
\\

COCO-Text
& Scene text detection / recognition
& \href{https://huggingface.co/datasets/howard-hou/COCO-Text}{Hugging Face}
\\

DocVQA
& Document understanding
& \href{https://huggingface.co/datasets/lmms-lab/DocVQA/tree/main/DocVQA}{Hugging Face}
\\

InfographicVQA
& Infographic understanding
& \href{https://huggingface.co/datasets/lmms-lab/DocVQA/tree/main/InfographicVQA}{Hugging Face}
\\

VisualMRC
& Text-rich image understanding
& \href{https://huggingface.co/datasets/NTT-hil-insight/VisualMRC}{Hugging Face}
\\

Kleister Charity
& Key information extraction
& \href{https://huggingface.co/datasets/orgrctera/kleister_charity_information_extraction}{Hugging Face}
\\

DeepForm
& Form understanding
& \href{https://github.com/jstray/deepform}{GitHub}
\\

ChartQA
& Chart understanding
& \href{https://huggingface.co/datasets/ahmed-masry/ChartQA}{Hugging Face}
\\

PubTabNet
& Table recognition
& \href{https://huggingface.co/datasets/ajimeno/PubTabNet}{Hugging Face}
\\

LaTeX-OCR
& Formula recognition
& \href{https://huggingface.co/datasets/lukbl/LaTeX-OCR-dataset}{Hugging Face}
\\

CASIA-HWDB
& Handwritten Chinese recognition
& \href{https://nlpr.ia.ac.cn/databases/handwriting/home.html}{Official}
\\

CORD
& Receipt understanding
& \href{https://huggingface.co/datasets/naver-clova-ix/cord-v2}{Hugging Face}
\\

SROIE
& Receipt OCR / information extraction
& \href{https://huggingface.co/datasets/darentang/sroie}{Hugging Face}
\\

\bottomrule
\end{tabular}

\end{table}

\subsection{Baseline Details}
\label{appbaselines}

Within each experimental setting, all controlled distillation methods share
the same student initialization, teacher configuration, training data,
optimization hyperparameters, and optimizer-step budget.
All distributional losses use the same teacher top-$64$ truncation,
unnormalized teacher probabilities, and zero-clamping implementation
described in Appendix~\ref{apptopk}.
The methods differ in rollout construction and, for CA-OPD, the supervision
assigned to replaced positions.

\paragraph{Offline KD.}
The teacher generates output sequences before training, and these sequences
remain fixed throughout distillation.
At each position, the student matches the teacher's next-token distribution
conditioned on the same fixed prefix.\citep{hinton2015distilling}.
This baseline therefore uses soft distributional targets rather than
hard token targets.

\paragraph{OPD.}
Following \citet{agarwal2024policy}, rollouts are generated online by the
current student policy.
The teacher provides next-token distributions conditioned on the
student-generated prefixes, and distributional supervision is applied at
every valid response position.
No student proposals are replaced during rollout construction.

\paragraph{SKD.}
We use the rank-based acceptance rule of \citet{xu2025speculative},
retaining a student proposal when its teacher rank is at most $K$,
with $K=25$ fixed throughout training.
Rejected proposals are replaced by the teacher argmax, and generation
continues from the resulting mixed prefix.
The same top-$64$ truncated forward-KL loss is applied at all valid
response positions, including replaced positions.

\section{Benchmark and Evaluation Details}
\label{appbenchmarks}

\paragraph{Common protocol.}
Within each benchmark column, every method uses an identical evaluation
harness, prompt, preprocessing pipeline, and decoding configuration. Reported
means and standard deviations are computed across the three independent
training runs; decoding-trial variation within a run is not mixed into the
run-level deviations.

\paragraph{GUI grounding.}
ScreenSpot-v2 and ScreenSpot-Pro use the official Qwen3.5 tool-call prompt, with
the assistant turn prefilled up to the coordinate field and images resized by
the official smart-resize rule. Each example is decoded eight times using the
official sampling configuration: temperature $0.7$, top-$p$ $0.8$, top-$k$
$20$, and presence penalty $1.5$. The score is the fraction of trials whose
predicted point lies inside the target box.

\paragraph{OCR.}
OCRBench-v2 is evaluated in English and Chinese with the official scorer.
CC-OCR is the macro average over its four tracks, and OmniDocBench v1.5 is
reported as full-page $1{-}\mathrm{NED}$. All OCR benchmarks use greedy
decoding with thinking disabled for every model.

\paragraph{Retention.}
RefCOCO, RefCOCO+, and RefCOCOg comprise eight official splits in total and are
evaluated with greedy decoding and ACC@0.5. Table~\ref{tabmain} reports their
mean over all eight splits, while Appendix~\ref{appmulti} reports each dataset
separately. MMBench uses EN-dev-v1.1 accuracy.

\section{Detailed Results and Run-Level Robustness}
\label{secres}

This section expands the main results in Table~\ref{tabmain} with
category-level target scores, dataset-level retention scores, and cross-run
standard deviations.

\subsection{Fine-Grained Target-Benchmark Results}
\label{appfinegrained}

\begin{table}[htbp]
\centering
\caption{
\textbf{ScreenSpot-Pro results by application category.}
All distillation results are means over three runs. Best and second-best
student results are shown in \textbf{bold} and \underline{underline},
respectively; the principal CA-OPD row is shaded.
}
\label{tabfinegrained}

\begingroup
\fontsize{7.5pt}{8.5pt}\selectfont
\renewcommand{\arraystretch}{1.08}
\setlength{\tabcolsep}{4pt}

\begin{tabular}{lccccccc}
\toprule

\rowcolor{tablegray}
Method & CAD & Creative & Dev & OS & Office & Scientific & Overall \\
\midrule

\rowcolor{tablegray}
Student (SFT-init)
& 29.69 & 34.42 & 33.53 & 40.05 & 54.62 & 41.14 & 38.19 \\
\noalign{\vskip-0.2pt}

\rowcolor{tablegray}
Offline KD
& \underline{40.01} & \underline{38.38} & \underline{38.94}
& \underline{43.75} & \underline{64.26} & \underline{44.91}
& \underline{44.23} \\
\noalign{\vskip-0.2pt}

\rowcolor{tablegray}
OPD
& 36.21 & 34.54 & 35.52 & 37.93 & 61.99 & 43.27 & 40.82 \\
\noalign{\vskip-0.2pt}

\rowcolor{tablegray}
SKD
& 38.06 & 34.03 & 35.72 & 39.29 & 60.51 & 44.49 & 41.20 \\
\noalign{\vskip-0.2pt}

\rowcolor{caopdblue}
\textbf{CA-OPD, annealed $\alpha:0.50\!\rightarrow\!0.25$}
& \textbf{41.63} & \textbf{39.41} & \textbf{41.11}
& \textbf{45.75} & \textbf{65.60} & \textbf{47.03}
& \textbf{45.92} \\

\bottomrule
\end{tabular}

\endgroup
\end{table}

\begin{table}[htbp]
\centering
\caption{
\textbf{OCRBench-v2 results by capability category.}
English and Chinese scores are macro-averaged where both are available;
Referring, Spotting, and Calculation are English-only. All distillation
results are means over three runs. Best and second-best student results are
shown in \textbf{bold} and \underline{underline}, respectively.
}
\label{tabfinegrainedocr}

\begingroup
\fontsize{7.5pt}{8.5pt}\selectfont
\renewcommand{\arraystretch}{1.08}
\setlength{\tabcolsep}{3pt}

\begin{tabular}{lcccccccc}
\toprule

\rowcolor{tablegray}
Method & Recog. & Referring & Spotting & Extract.
& Parsing & Calc. & Underst. & Reason. \\
\midrule

\rowcolor{tablegray}
Student (SFT-init)
& \textbf{70.06} & 41.06 & 18.30 & 72.27
& 47.09 & \textbf{38.03} & \underline{54.99} & \textbf{39.07} \\
\noalign{\vskip-0.2pt}

\rowcolor{tablegray}
Offline KD
& 69.02 & 46.93 & \underline{22.34} & 75.48
& \underline{49.21} & 36.33 & 54.46 & 37.55 \\
\noalign{\vskip-0.2pt}

\rowcolor{tablegray}
OPD
& 67.95 & 46.96 & 19.42 & 74.30
& 48.87 & 36.05 & 54.51 & 37.43 \\
\noalign{\vskip-0.2pt}

\rowcolor{tablegray}
SKD
& 68.16 & \underline{47.63} & 22.17 & \underline{75.66}
& 47.59 & 35.10 & 54.28 & \underline{38.14} \\
\noalign{\vskip-0.2pt}

\rowcolor{caopdblue}
\textbf{CA-OPD, annealed $\alpha:0.50\!\rightarrow\!0.25$}
& \underline{69.42} & \textbf{48.32} & \textbf{24.63}
& \textbf{76.16} & \textbf{49.30} & \underline{36.59}
& \textbf{55.29} & 38.00 \\

\bottomrule
\end{tabular}

\endgroup
\end{table}

Tables~\ref{tabfinegrained} and~\ref{tabfinegrainedocr} show that the gains are not driven by a single subset. 
CA-OPD is best in all six ScreenSpot-Pro application categories. 
On OCRBench-v2, it exceeds OPD in all eight capability categories and attains the best listed score in five; the SFT initialization
remains strongest in Recognition, Calculation, and Reasoning.

\subsection{Held-Out Capability Retention}
\label{appmulti}

Table~\ref{tabretention} decomposes the aggregate RefCOCO entry in
Table~\ref{tabmain}. Across the controlled distillation methods, the spread is
at most $0.43$ points in any column, and no method moves a dataset materially
away from the SFT initialization. Thus, the target-task gains are not obtained
at the expense of the held-out capabilities measured here.

\begin{table}[htbp]
\centering
\caption{
\textbf{Retention results by dataset.}
RefCOCO/+/g are ACC@0.5 averaged over each dataset's official splits; MMBench
is EN-dev-v1.1 accuracy. Values are means over three runs. Best and second-best
student results are shown in \textbf{bold} and \underline{underline}.
}
\label{tabretention}

\begingroup
\fontsize{7.5pt}{8.5pt}\selectfont
\renewcommand{\arraystretch}{1.08}
\setlength{\tabcolsep}{4.5pt}

\begin{tabular}{lcccc}
\toprule

\rowcolor{tablegray}
Method & RefCOCO & RefCOCO+ & RefCOCOg & MMBench \\
\midrule

\rowcolor{tablegray}
Student (SFT-init)
& \textbf{85.63}
& \textbf{78.08}
& 82.44
& 79.16 \\
\noalign{\vskip-0.2pt}

\rowcolor{tablegray}
Offline KD
& 85.49
& 77.71
& \underline{82.62}
& 79.99 \\
\noalign{\vskip-0.2pt}

\rowcolor{tablegray}
OPD
& 85.38
& 77.66
& 82.54
& \underline{80.00} \\
\noalign{\vskip-0.2pt}

\rowcolor{tablegray}
SKD
& 85.43
& \underline{77.77}
& 82.44
& 79.76 \\
\noalign{\vskip-0.2pt}

\rowcolor{tablegray}
CA-OPD, fixed $\alpha=0.25$
& \underline{85.54}
& 77.60
& 82.55
& 79.73 \\
\noalign{\vskip-0.2pt}

\rowcolor{caopdblue}
\textbf{CA-OPD, annealed $\alpha:0.50\!\rightarrow\!0.25$}
& 85.53
& \underline{77.77}
& \textbf{82.64}
& \textbf{80.16} \\

\bottomrule
\end{tabular}

\endgroup
\end{table}

\subsection{Cross-Run Variability}
\label{appseeds}

Table~\ref{tabrunvar} reports standard deviations across the three independent
runs behind each mean in Table~\ref{tabmain}. Runs share all hyperparameters
and differ in data order and stochastic on-policy sampling, so the table
captures run-to-run variability rather than decoding-trial variance. Ungated
OPD is notably unstable on SS-Pro, CC-OCR, and OmniDoc; the annealed gate lowers
those deviations from $1.46$, $0.96$, and $1.13$ to $0.72$, $0.51$, and $0.33$,
respectively.

\begin{table}[htbp]
\centering
\caption{
\textbf{Cross-run standard deviation.}
Lower is better. Best and second-best values are shown in \textbf{bold} and
\underline{underline}, respectively.
}
\label{tabrunvar}

\begingroup
\fontsize{7.5pt}{8.5pt}\selectfont
\renewcommand{\arraystretch}{1.08}
\setlength{\tabcolsep}{3pt}

\begin{tabular}{lcccccccc}
\toprule

\rowcolor{tablegray}
Method & SS-v2 & SS-Pro & OCRv2-EN & OCRv2-ZH
& CC-OCR & OmniDoc & RefCOCO & MMBench \\
\midrule

\rowcolor{tablegray}
Offline KD
& 0.29
& \textbf{0.40}
& 0.57
& \underline{0.17}
& \underline{0.33}
& \textbf{0.07}
& \underline{0.07}
& 0.18 \\
\noalign{\vskip-0.2pt}

\rowcolor{tablegray}
OPD
& 0.74
& 1.46
& \underline{0.25}
& 0.59
& 0.96
& 1.13
& 0.13
& \underline{0.06} \\
\noalign{\vskip-0.2pt}

\rowcolor{tablegray}
SKD
& \underline{0.25}
& \underline{0.41}
& 0.71
& 0.35
& \textbf{0.30}
& 0.48
& \textbf{0.04}
& 0.12 \\
\noalign{\vskip-0.2pt}

\rowcolor{tablegray}
CA-OPD, fixed $\alpha=0.25$
& 0.47
& 0.53
& \textbf{0.17}
& 0.30
& 0.38
& \underline{0.23}
& 0.15
& 0.20 \\
\noalign{\vskip-0.2pt}

\rowcolor{caopdblue}
\textbf{CA-OPD, annealed $\alpha:0.50\!\rightarrow\!0.25$}
& \textbf{0.18}
& 0.72
& 0.41
& \textbf{0.10}
& 0.51
& 0.33
& 0.14
& \textbf{0.05} \\

\bottomrule
\end{tabular}

\endgroup
\end{table}

\section{Intervention Budget, Schedule, and Gating Dynamics}
\label{appgatinganalysis}

\subsection{Full Intervention Budget Sweep}
\label{appbudgetsweep}

Table~\ref{tabbudgetfull} lists the operating points behind
Figure~\ref{figgatinganalysis}(a). The sweep shows that accuracy does not increase
monotonically with replacement rate: random placement offers little benefit at
a comparable budget, and high-frequency fixed floors do not consistently beat
the annealed configuration. This supports the main-text conclusion that the
location and timing of interventions matter in addition to their frequency.

\begin{table}[htbp]
\centering
\caption{
\textbf{Full intervention-budget sweep.}
Replace rate is the median fraction of rollout tokens replaced during
training. Values are means over three runs except rows marked \(\dagger\),
which are single runs. Bold and underlined values rank only the multi-run
configurations.
}
\label{tabbudgetfull}

\begingroup
\fontsize{7.5pt}{8.5pt}\selectfont
\renewcommand{\arraystretch}{1.08}
\setlength{\tabcolsep}{4.5pt}

\begin{tabular}{lccc}
\toprule

\rowcolor{tablegray}
Variant & Replace rate (\%) & SS-Pro & OCRv2-EN \\
\midrule

\rowcolor{tablegray}
OPD (no gate)
& 0.00
& 40.82
& 50.79 \\
\noalign{\vskip-0.2pt}

\rowcolor{tablegray}
Rank criterion (SKD, $K{=}25$)
& 0.16
& 41.20
& 51.30 \\
\noalign{\vskip-0.2pt}

\rowcolor{tablegray}
Rank criterion, closest budget ($K{=}1$)
& 3.41
& 44.00
& 51.61 \\
\noalign{\vskip-0.2pt}

\rowcolor{tablegray}
Random placement\textsuperscript{\(\dagger\)}
& 3.87
& 40.20
& 50.62 \\
\noalign{\vskip-0.2pt}

\rowcolor{tablegray}
Fixed floor $\alpha=0.10$
& 4.0
& 44.19
& 52.29 \\
\noalign{\vskip-0.2pt}

\rowcolor{tablegray}
Fixed floor $\alpha=0.25$
& 4.9
& 44.58
& \underline{52.44} \\
\noalign{\vskip-0.2pt}

\rowcolor{tablegray}
Fixed floor $\alpha=0.50$
& 6.0
& \underline{45.43}
& 52.22 \\
\noalign{\vskip-0.2pt}

\rowcolor{tablegray}
Fixed floor $\alpha=0.90$\textsuperscript{\(\dagger\)}
& 7.8
& 45.85
& 52.54 \\
\noalign{\vskip-0.2pt}

\rowcolor{tablegray}
Fixed floor $\alpha=0.98$\textsuperscript{\(\dagger\)}
& 9.6
& 45.88
& 52.06 \\
\noalign{\vskip-0.2pt}

\rowcolor{caopdblue}
\textbf{CA-OPD, annealed $\alpha:0.50\!\rightarrow\!0.25$}
& 5.4
& \textbf{45.92}
& \textbf{52.47} \\

\bottomrule
\end{tabular}

\endgroup
\end{table}

\subsection{Annealing-Schedule Variants}
\label{appschedulevariants}

\paragraph{Threshold selection.}
We specified $\alpha_{\mathrm{start}}=0.50$ and
$\alpha_{\mathrm{end}}=0.25$ as the default probability-floor endpoints
before conducting the sensitivity experiments.
The fixed-threshold ablation uses $\alpha=0.25$, matching the terminal
floor of the main schedule.
Subsequent experiments vary the threshold values, schedule shape, and
direction to assess sensitivity to these choices.

Table~\ref{tabschedulevariants} isolates the schedule. Both direction and
terminal floor matter: relaxing to $\alpha=0.10$ loses part of the gain,
linear relaxation is weaker than cosine relaxation at the same endpoints, and
reversing the schedule loses more. An earlier and stricter start at
$\alpha_{\mathrm{start}}=0.70$ is included as a single-run sensitivity check.

\begin{table}[htbp]
\centering
\caption{
\textbf{Annealing-schedule variants.}
All rows share the confidence criterion, argmax writeback, and
intervention-aligned supervision of the main method. Values are means over
three runs except \(\dagger\). Bold and underlined values rank only multi-run
configurations.
}
\label{tabschedulevariants}

\begingroup
\fontsize{7.5pt}{8.5pt}\selectfont
\renewcommand{\arraystretch}{1.08}
\setlength{\tabcolsep}{4.5pt}

\begin{tabular}{lcc}
\toprule

\rowcolor{tablegray}
Schedule & SS-Pro & OCRv2-EN \\
\midrule

\rowcolor{caopdblue}
\textbf{Cosine $\alpha:0.50\!\rightarrow\!0.25$ (main)}
& \textbf{45.92}
& \textbf{52.47} \\
\noalign{\vskip-0.2pt}

\rowcolor{tablegray}
Cosine $\alpha:0.50\!\rightarrow\!0.10$
& \underline{45.06}
& \underline{52.27} \\
\noalign{\vskip-0.2pt}

\rowcolor{tablegray}
Linear $\alpha:0.50\!\rightarrow\!0.10$
& 44.54
& 51.61 \\
\noalign{\vskip-0.2pt}

\rowcolor{tablegray}
Reversed cosine $\alpha:0.10\!\rightarrow\!0.50$
& 44.61
& 51.49 \\
\noalign{\vskip-0.2pt}

\rowcolor{tablegray}
Reversed cosine $\alpha:0.25\!\rightarrow\!0.50$
& 44.36
& 51.69 \\
\noalign{\vskip-0.2pt}

\rowcolor{tablegray}
Cosine $\alpha:0.70\!\rightarrow\!0.10$\textsuperscript{\(\dagger\)}
& 45.37
& 52.24 \\

\bottomrule
\end{tabular}

\endgroup
\end{table}
\subsection{Gating Dynamics over Training}
\label{appgatingdynamics}

\begin{figure}[htbp]
\centering
\IfFileExists{images/dynamics_v2.pdf}{
\includegraphics[width=\linewidth]{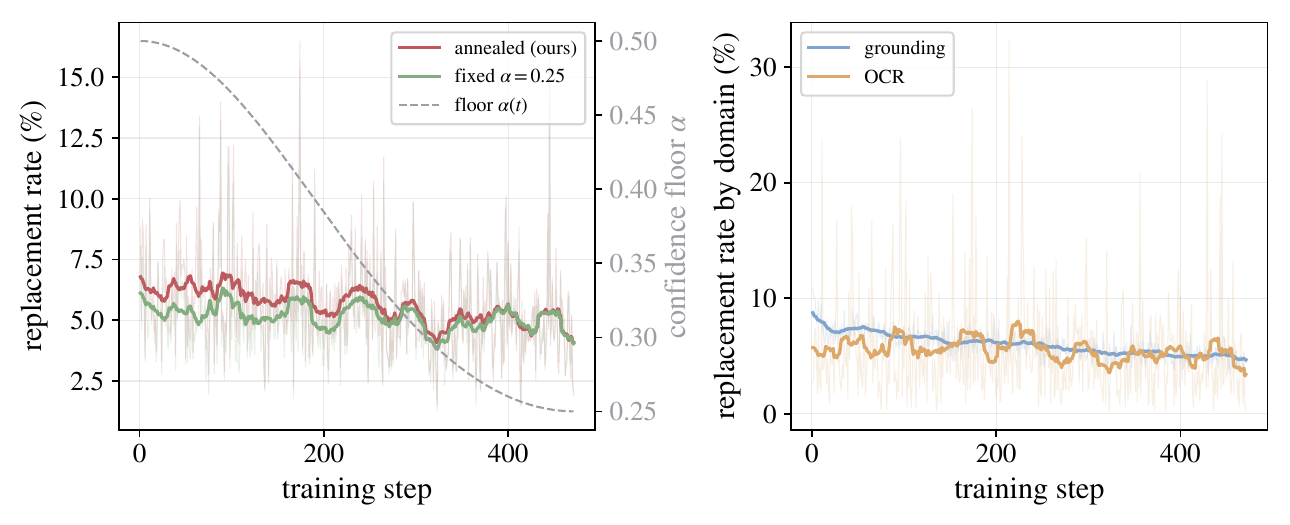}
}{
\placeholderbox{4.4cm}{\todofig{dynamics figure missing}}
}
\caption{
\textbf{Gating dynamics over training.}
Replacement rate decreases with optimization. For the annealed schedule, the
trend reflects both threshold relaxation and increasing student reliability;
the fixed-threshold curve isolates the latter effect.
}
\label{figgatingdynamics}
\end{figure}

Figure~\ref{figgatingdynamics} reports the replacement rate throughout
training for the annealed and fixed-threshold variants. Under annealing,
intervention is more frequent early and decreases as the criterion relaxes.
The fixed-threshold run also declines, showing that increasing student
reliability independently reduces the number of proposals requiring teacher
correction.

\subsection{Counterfactual Outcome Counts}
\label{appcounterfactual}

Table~\ref{tabcounterfactualcounts} reports the complete outcome counts
for Figure~\ref{figgatinganalysis}(c).
The gate-triggered and gate-passed groups each contain $95$ grounding
pairs and $64$ OCR pairs.

\begin{table}[htbp]
\centering
\caption{
\textbf{Outcome counts for paired counterfactual comparisons.}
Better, Worse, and Unchanged denote positive, negative, and zero
differences in the outcome measures plotted in
Figure~\ref{figgatinganalysis}(c), respectively.
$N$ includes all three outcomes.
}
\label{tabcounterfactualcounts}

\begingroup
\fontsize{7.5pt}{8.5pt}\selectfont
\renewcommand{\arraystretch}{1.12}
\setlength{\tabcolsep}{5pt}

\begin{tabular}{llrrrrr}
\toprule
\rowcolor{tablegray}
Domain & Group & $N$ & Better & Worse
& Unchanged & Unchanged (\%) \\
\midrule

Grounding & Gate-triggered
& 95 & 58 & 21 & 16 & 16.8 \\

          & Gate-passed
& 95 & 4 & 2 & 89 & 93.7 \\

\midrule

OCR & Gate-triggered
& 64 & 33 & 25 & 6 & 9.4 \\

    & Gate-passed
& 64 & 1 & 1 & 62 & 96.9 \\

\bottomrule
\end{tabular}

\endgroup
\end{table}

\section{Stratified Protocols and Results}
\label{appstratified}

This section supplies the bucket definitions used in
Section~\ref{secstratifiedmain} before presenting the corresponding detailed
results. All buckets are defined from the SFT initialization before any
compared method is trained, so bucket membership is a pre-treatment covariate.

\subsection{Stratification Protocols}
\label{appstratprotocol}

\paragraph{Output length.}
OCRBench-v2 examples, with English and Chinese pooled, are sorted by reference
transcription length in characters and split into exact tertiles. Each cell in
the output-length row of Table~\ref{tabstratifiedmain} reports the within-bucket
mean of the official per-example scores, averaged over three training runs.
Because task types are pooled, only within-bucket differences between methods
are meaningful; the corresponding CA-OPD-over-OPD deltas are therefore
summarized directly in the main table rather than repeated as absolute scores
here.

\paragraph{Initial competence on SS-Pro.}
The eight evaluation rollouts of the SFT-initialized student are split in half:
the first four define the bucket (all wrong, mixed, or all correct), and every
reported score, including the SFT-init row, uses only the remaining four. This
split avoids defining and evaluating the buckets on the same trials, which
would otherwise force the all-correct initialization score to $100$ and inflate
selection effects.

\paragraph{Within-type OCR difficulty.}
OCRBench-v2 uses a single greedy output, so difficulty tertiles are computed
within each task type by sorting examples on the SFT-init score. Within-type
bucketing is necessary because continuous metrics place partially correct
examples in the middle of the range, whereas binary metrics do not; a global
split would therefore confound difficulty with metric type and output length.

\paragraph{Teacher-rejection density.}
For each SS-Pro example, the SFT-init greedy output is teacher-force scored by
the grounding teacher. We define $\rho$ as the fraction of output tokens whose
teacher probability is below $0.25$, the terminal floor of the main annealed
schedule, and split examples into tertiles. Thus, low $\rho$ corresponds to
strong teacher endorsement and high $\rho$ to frequent rejection. The tercile
ordering and the flat CA-OPD-over-OPD margin are unchanged at thresholds $0.1$
and $0.5$. This statistic correlates with initial competence ($r=-0.51$) but
not with output length ($r=-0.10$).

\subsection{SS-Pro by Initial Student Competence}
\label{appinitialcompetence}

Table~\ref{tabinitcompetencefull} shows that CA-OPD improves on OPD in every
initial-competence group. It also better preserves behavior already mastered by
the SFT initialization: in the all-correct group, OPD falls from $96.03$ to
$90.18$, whereas annealed CA-OPD retains $93.83$.

\begin{table}[htbp]
\centering
\caption{
\textbf{SS-Pro performance by initial student competence.}
Examples are grouped using four held-out SFT-init rollouts and evaluated using
the remaining four. Bold and underlined values denote the best and second-best
controlled distillation results; SFT-init is a reference row.
}
\label{tabinitcompetencefull}

\begingroup
\fontsize{7.5pt}{8.5pt}\selectfont
\renewcommand{\arraystretch}{1.08}
\setlength{\tabcolsep}{4.5pt}

\begin{tabular}{lccc}
\toprule

\rowcolor{tablegray}
Method
& \makecell{All wrong\\($n{=}822$)}
& \makecell{Mixed\\($n{=}312$)}
& \makecell{All correct\\($n{=}447$)} \\
\midrule

\rowcolor{tablegray}
Student (SFT-init)
& 2.49
& 49.44
& 96.03 \\
\noalign{\vskip-0.2pt}

\rowcolor{tablegray}
OPD
& 9.88
& 51.47
& 90.18 \\
\noalign{\vskip-0.2pt}

\rowcolor{tablegray}
CA-OPD, fixed $\alpha=0.25$
& \underline{13.39}
& \underline{58.60}
& \underline{92.04} \\
\noalign{\vskip-0.2pt}

\rowcolor{caopdblue}
\textbf{CA-OPD, annealed $\alpha:0.50\!\rightarrow\!0.25$}
& \textbf{13.79}
& \textbf{62.07}
& \textbf{93.83} \\

\bottomrule
\end{tabular}

\endgroup
\end{table}

\subsection{OCRBench-v2 by Within-Type Difficulty}
\label{appocrdifficulty}

Table~\ref{tabstratocr} reports bucket-level micro means. The annealed method
improves over OPD in every stratum of both languages; the fixed floor falls
below OPD only in the hardest Chinese tertile. The extreme SFT-init values are
a consequence of defining the buckets on that same greedy output and should be
read as a reference rather than a controlled comparison.

\begin{table}[htbp]
\centering
\caption{
\textbf{OCRBench-v2 by within-type difficulty tertiles.}
Values are bucket-level micro means and are not directly comparable with the
macro-averaged headline scores. Bold and underlined values denote the best and
second-best controlled distillation results; SFT-init is a reference row.
}
\label{tabstratocr}

\begingroup
\fontsize{7.5pt}{8.5pt}\selectfont
\renewcommand{\arraystretch}{1.08}
\setlength{\tabcolsep}{3.5pt}

\begin{tabular}{lcccccc}
\toprule

\rowcolor{tablegray}
& \multicolumn{3}{c}{OCRv2-EN}
& \multicolumn{3}{c}{OCRv2-ZH} \\

\cmidrule(lr){2-4}\cmidrule(lr){5-7}

\rowcolor{tablegray}
Method & Hard & Medium & Easy & Hard & Medium & Easy \\
\midrule

\rowcolor{tablegray}
Student (SFT-init)
& 18.04
& 59.74
& 90.52
& 15.91
& 45.05
& 78.86 \\
\noalign{\vskip-0.2pt}

\rowcolor{tablegray}
OPD
& 26.26
& 58.34
& 81.65
& \underline{25.82}
& 45.39
& 72.43 \\
\noalign{\vskip-0.2pt}

\rowcolor{tablegray}
CA-OPD, fixed $\alpha=0.25$
& \underline{26.63}
& \underline{59.92}
& \underline{82.70}
& 25.49
& \underline{45.90}
& \underline{73.36} \\
\noalign{\vskip-0.2pt}

\rowcolor{caopdblue}
\textbf{CA-OPD, annealed $\alpha:0.50\!\rightarrow\!0.25$}
& \textbf{27.00}
& \textbf{60.10}
& \textbf{82.77}
& \textbf{26.26}
& \textbf{46.32}
& \textbf{73.60} \\

\bottomrule
\end{tabular}

\endgroup
\end{table}

\subsection{SS-Pro by Teacher-Rejection Density}
\label{apprejectiondensity}

Table~\ref{tabstratrejection} shows an almost constant annealed-CA-OPD margin
over OPD across rejection-density tertiles: $+4.95$, $+5.31$, and $+5.05$.
In the low-$\rho$ group, where the initialization is most strongly endorsed by
the teacher, OPD falls below the initialization while both gated variants
improve on it.

\begin{table}[htbp]
\centering
\caption{
\textbf{SS-Pro by teacher-rejection-density tertiles.}
$\rho$ is the fraction of SFT-init output tokens with teacher probability below
$0.25$; each tertile contains $527$ examples. Scores average eight decoding
trials and three training runs. Bold and underlined values denote the best and
second-best controlled distillation results; SFT-init is a reference row.
}
\label{tabstratendorse}
\label{tabstratrejection}

\begingroup
\fontsize{7.5pt}{8.5pt}\selectfont
\renewcommand{\arraystretch}{1.08}
\setlength{\tabcolsep}{4.5pt}

\begin{tabular}{lccc}
\toprule

\rowcolor{tablegray}
Method & Low $\rho$ & Medium $\rho$ & High $\rho$ \\
\midrule

\rowcolor{tablegray}
Student (SFT-init)
& 65.37
& 37.83
& 11.36 \\
\noalign{\vskip-0.2pt}

\rowcolor{tablegray}
OPD
& 63.87
& 40.23
& 18.35 \\
\noalign{\vskip-0.2pt}

\rowcolor{tablegray}
CA-OPD, fixed $\alpha=0.25$
& \underline{67.61}
& \underline{44.25}
& \underline{21.88} \\
\noalign{\vskip-0.2pt}

\rowcolor{caopdblue}
\textbf{CA-OPD, annealed $\alpha:0.50\!\rightarrow\!0.25$}
& \textbf{68.82}
& \textbf{45.53}
& \textbf{23.40} \\

\midrule

\rowcolor{tablegray}
$\Delta$ (annealed $-$ OPD)
& $+4.95$
& $+5.31$
& $+5.05$ \\

\bottomrule
\end{tabular}

\endgroup
\end{table}

\section{Additional Comparisons}
\label{appadditionalcomparisons}

\subsection{Additional Evaluation with Qwen3-VL}
\label{appqwen3vl}

We conduct an additional evaluation using Qwen3-VL-8B-Instruct as the
frozen teacher and Qwen3-VL-2B-Instruct as the student.
Both models are initialized directly from their released Instruct checkpoints,
without additional task-specific supervised fine-tuning before distillation.
All compared methods use the same Qwen3-VL-2B-Instruct initialization,
distillation data, optimization budget, and benchmark evaluation protocols.
The training data composition, decoding configurations, and evaluation metrics
are aligned with those used in the main Qwen3.5 experiments.

\begin{table*}[t]
\centering
\caption{
\textbf{Additional evaluation with Qwen3-VL.}
Results for the distillation methods are reported as mean $\pm$ standard deviation over three independent runs, while the undistilled student and teacher are single-checkpoint reference points.
Best and second-best results among all student-side configurations,
including the undistilled student, are shown in \textbf{bold} and
\underline{underline}, respectively, the teacher is excluded from ranking.
}
\label{tabqwen3vl}

\begingroup
\renewcommand{\arraystretch}{1.08}
\setlength{\tabcolsep}{3.5pt}

\resizebox{\textwidth}{!}{%
\begin{tabular}{lcccccccc}
\toprule

\rowcolor{tablegray}
&
\multicolumn{2}{c}{GUI Grounding}
& \multicolumn{4}{c}{OCR}
& \multicolumn{2}{c}{Retention} \\
\noalign{\vskip -3pt}
\cmidrule(lr){2-3}
\cmidrule(lr){4-7}
\cmidrule(lr){8-9}
\noalign{\vskip -3pt}

\rowcolor{tablegray}
Method
& SS-v2
& SS-Pro
& OCRv2-EN
& OCRv2-ZH
& CC-OCR
& OmniDoc
& RefCOCO
& MMBench \\
\midrule

\rowcolor{tablegray}
Teacher (Qwen3-VL-8B-Instruct)
& 94.94
& 56.20
& 64.57
& 62.36
& 76.38
& 68.66
& 88.33
& 89.60 \\

\rowcolor{tablegray}
Student (Qwen3-VL-2B-Instruct)
& 88.77
& 41.14
& 55.61
& 54.42
& 64.61
& 64.90
& \textbf{84.30}
& 84.06 \\

\midrule

\rowcolor{tablegray}
Offline KD~\citep{hinton2015distilling}
& \underline{91.08}\,{\scriptsize$\pm$0.20}
& \textbf{48.64}\,{\scriptsize$\pm$0.47}
& \underline{57.17}\,{\scriptsize$\pm$0.38}
& \underline{54.78}\,{\scriptsize$\pm$0.27}
& \textbf{66.16}\,{\scriptsize$\pm$0.29}
& 65.95\,{\scriptsize$\pm$0.59}
& 84.00\,{\scriptsize$\pm$0.14}
& 84.05\,{\scriptsize$\pm$0.20} \\

\rowcolor{tablegray}
OPD~\citep{agarwal2024policy}
& 90.86\,{\scriptsize$\pm$0.50}
& 46.40\,{\scriptsize$\pm$0.85}
& 56.77\,{\scriptsize$\pm$0.28}
& 54.43\,{\scriptsize$\pm$0.40}
& 64.89\,{\scriptsize$\pm$0.07}
& 65.59\,{\scriptsize$\pm$0.67}
& \underline{84.05}\,{\scriptsize$\pm$0.15}
& 84.00\,{\scriptsize$\pm$0.13} \\

\rowcolor{tablegray}
SKD~\citep{xu2025speculative}
& 90.93\,{\scriptsize$\pm$0.22}
& 46.58\,{\scriptsize$\pm$0.78}
& 56.65\,{\scriptsize$\pm$0.32}
& \textbf{54.82}\,{\scriptsize$\pm$0.54}
& 65.01\,{\scriptsize$\pm$0.55}
& \underline{66.28}\,{\scriptsize$\pm$2.33}
& 84.04\,{\scriptsize$\pm$0.18}
& \underline{84.08}\,{\scriptsize$\pm$0.00} \\

\rowcolor{caopdblue}
\textbf{CA-OPD, annealed $\alpha:0.50\!\rightarrow\!0.25$}
& \textbf{91.35}\,{\scriptsize$\pm$0.27}
& \underline{48.36}\,{\scriptsize$\pm$0.50}
& \textbf{57.18}\,{\scriptsize$\pm$0.31}
& 54.32\,{\scriptsize$\pm$0.41}
& \underline{65.87}\,{\scriptsize$\pm$0.57}
& \textbf{67.01}\,{\scriptsize$\pm$2.08}
& 83.82\,{\scriptsize$\pm$0.14}
& \textbf{84.27}\,{\scriptsize$\pm$0.12} \\

\bottomrule
\end{tabular}%
}

\endgroup
\end{table*}

It achieves the strongest results among the compared distillation methods
on ScreenSpot-v2, OCRBench-v2 English, OmniDocBench, and MMBench.
These results indicate that the benefits of CA-OPD are not restricted to
the SFT-initialized Qwen3.5 setting.

\subsection{Single-Teacher Grounding Comparison}
\label{appsingleteacher}

Table~\ref{tabsingleteacher} checks that the benefit of confidence gating does
not depend on multi-teacher routing or on the SFT initialization used in the
main experiments. Here, the student is the raw pretrained Qwen3.5-0.8B, a
single grounding teacher supervises one domain, and the gate uses a fixed floor
$\alpha=0.25$ for $699$ optimizer steps. Because this is a single-run setting
with a different initialization and training regime, its absolute values are
not comparable to Table~\ref{tabmain}; only the within-table OPD--CA-OPD
comparison is meaningful. CA-OPD improves SS-v2 by $1.74$ points and SS-Pro by
$5.48$ points based on the displayed values.

\begin{table}[htbp]
\centering
\caption{
\textbf{Single-teacher grounding comparison.}
The student is raw Qwen3.5-0.8B, the gate uses fixed $\alpha=0.25$, and results
come from one run. Grounding follows the main-table protocol of eight sampled
trials.
}
\label{tabsingleteacher}

\begingroup
\fontsize{7.5pt}{8.5pt}\selectfont
\renewcommand{\arraystretch}{1.08}
\setlength{\tabcolsep}{4.5pt}

\begin{tabular}{lcc}
\toprule

\rowcolor{tablegray}
Method & SS-v2 & SS-Pro \\
\midrule

\rowcolor{tablegray}
OPD
& 87.67
& 40.64 \\
\noalign{\vskip-0.2pt}

\rowcolor{caopdblue}
\textbf{CA-OPD, fixed $\alpha=0.25$}
& \textbf{89.41}
& \textbf{46.12} \\

\bottomrule
\end{tabular}

\endgroup
\end{table}

\subsection{Policy-Gradient Baseline}
\label{apppg}

For completeness, we also trained a policy-gradient variant using a reverse-KL objective with a REINFORCE-style estimator under the same initialization, data, and optimizer budget as the main comparison. 
In one run it reaches $88.61$ on SS-v2, $45.56$ on SS-Pro, $51.58$ on OCRv2-EN, and $53.04$ on OCRv2-ZH. 
Its SS-Pro score is close to CA-OPD, but it optimizes a different objective from the teacher-distribution matching studied in the main paper and was not repeated across runs. 
We therefore keep it outside the controlled main table.

\section{Training Cost and Compute Matching}
\label{appcompute}

All methods run on the same $8\times$A800 node with identical data, batch size, and $471$ optimizer steps. 
Table~\ref{tabcompute} reports median wall-clock seconds per optimizer step over a full run. 
Both CA-OPD and SKD cost about $1.5\times$ ungated OPD because they perform interleaved teacher queries during generation; the gate arithmetic itself is not the dominant overhead.

\begin{table}[htbp]
\centering
\caption{
\textbf{Measured training cost.}
Values are median seconds per optimizer step under matched hardware, data, and
step count. Offline KD uses fixed teacher prefixes and therefore has no rollout
replacement rate.
}
\label{tabcompute}

\begingroup
\fontsize{7.5pt}{8.5pt}\selectfont
\renewcommand{\arraystretch}{1.08}
\setlength{\tabcolsep}{4.5pt}

\begin{tabular}{lccc}
\toprule

\rowcolor{tablegray}
Method & Seconds/step & Relative to OPD & Median replace rate (\%) \\
\midrule

\rowcolor{tablegray}
Offline KD
& 35.3
& 0.89
& \ph \\
\noalign{\vskip-0.2pt}

\rowcolor{tablegray}
OPD
& 39.6
& 1.00
& 0.00 \\
\noalign{\vskip-0.2pt}

\rowcolor{tablegray}
SKD ($K{=}25$)
& 58.6
& 1.48
& 0.16 \\
\noalign{\vskip-0.2pt}

\rowcolor{tablegray}
CA-OPD, fixed $\alpha=0.25$
& 58.3
& 1.47
& 4.9 \\
\noalign{\vskip-0.2pt}

\rowcolor{caopdblue}
\textbf{CA-OPD, annealed $\alpha:0.50\!\rightarrow\!0.25$}
& 59.3
& 1.50
& 5.4 \\

\bottomrule
\end{tabular}

\endgroup
\end{table}

The fixed-$\alpha$ and annealed-$\alpha$ throughput is taken from the run using the final chunked
gate implementation. Two earlier runs of that configuration used a per-token
implementation at approximately $109$ seconds per step; this changes
throughput, not the reported final scores, whose SS-Pro cross-run deviation is
$0.53$.
The reported timings reflect our current chunked implementation;
further systems-level optimization of interleaved rollouts is left
for future work.

\section{Limitations and Scope}
\label{applimitations}

CA-OPD assumes that teacher probabilities are sufficiently calibrated to
identify unreliable student transitions; systematic teacher--student mismatch,
such as domain shift, may require retuning the confidence floor. We validate
structured visual prediction with shared teacher--student tokenizers, leaving
mismatched tokenizers and open-ended generation untested.

Mixed-prefix rollout remains sequential in chunks and queries the teacher at
every visited state, producing about $1.5\times$ the wall-clock cost of
ungated OPD (Appendix~\ref{appcompute}). Gating also reduces rather than
eliminates erosion of mastered behavior: on consistently solved SS-Pro
examples, annealed CA-OPD reaches $93.83$ versus the initialization's $96.03$.
\end{document}